%% file: main.tex
\documentclass{article} %
\usepackage{arxiv,times}

\input{math_commands.tex}

\usepackage{hyperref}
\usepackage{url}
\usepackage{graphicx}
\usepackage{caption}
\usepackage{booktabs}
\usepackage{multirow}
\usepackage{amsmath,amssymb}
\usepackage{xcolor}
\usepackage{colortbl}         %
\usepackage{array}            %
\usepackage{adjustbox}        %
\usepackage{xspace}
\usepackage{enumitem}

\definecolor{grpgreen}{RGB}{223,240,216}
\definecolor{grpyellow}{RGB}{252,248,227}
\definecolor{grpred}{RGB}{242,222,222}
\newcolumntype{R}[2]{>{\adjustbox{angle=#1,lap=\width-(#2)}\bgroup}l<{\egroup}}

\graphicspath{{media/}{media/ppt_merge/}{src/figures/}{logo/}}

\title{SeekJudge: A Practical Reward Framework for Reinforcement Learning in Computer-Use Agents}

\author{Yang Wan$^{1}$, Zhenhao Zhang$^{2}$, Jierui Wang$^{1}$, Xiangxin Zhou$^{3}$, Linchao Zhu$^{1}$\thanks{Corresponding author.} \\
$^{1}$Zhejiang University \quad $^{2}$University of Rochester \quad $^{3}$Tencent
}

\newcommand{\cmark}{\textcolor{green!55!black}{\checkmark}}
\newcommand{\xmark}{\textcolor{red!70!black}{\ensuremath{\times}}}

\begin{document}

\maketitle

\begin{abstract}
\input{src/abstract}
\end{abstract}

\input{src/introduction}

\input{src/related_work}

\input{src/method}
\input{src/experiments}

\input{src/conclusion}

\section{Reproducibility Statement}

Section~\ref{sec:datagen} specifies the teacher models and the two prompt searches that build the distillation data, and Section~\ref{sec:bench} describes how the tasks of CUAStepBench are selected and labeled. The prompts that drive exploration and scoring, including the criteria prompt that produces the dense labels, are described in Appendix~\ref{app:context}. Appendix~\ref{app:training} lists the RL hyperparameters and the evaluation protocol, and Appendix~\ref{app:reward} gives the rule that combines the per-step scores into the trajectory reward. Appendix~\ref{app:cost} gives the token prices, the closed-source rates, and the prefix-cache rule that we apply uniformly to every method when computing inference cost. Appendix~\ref{app:leakage} describes the leakage control for the distillation data, the prompt search, and the score regression. We will release the training data, the SeekJudge-9B model, CUAStepBench, and the code of SeekJudge and its \mbox{reward server}.

\bibliography{refs}
\bibliographystyle{arxiv}

\clearpage
\appendix

\input{src/appendix}

\end{document}

%% file: math_commands.tex
\usepackage{amsmath,amsfonts,bm}

\def\eqref#1{equation~\ref{#1}}

\def\1{\bm{1}}

\DeclareMathAlphabet{\mathsfit}{\encodingdefault}{\sfdefault}{m}{sl}
\SetMathAlphabet{\mathsfit}{bold}{\encodingdefault}{\sfdefault}{bx}{n}

%% file: src/abstract.tex
Deciding whether a trajectory actually fulfills its instruction determines how we measure computer-use agents on long-horizon graphical-user-interface tasks and how we train them with reinforcement learning (RL).
This judgment has long relied on rule-based evaluation.
Rule-based evaluation often disagrees with human intention and becomes outdated when an app updates or its online content drifts.
Existing model-based judges attempt to address these problems, but their judging accuracy remains limited.
We propose the \textbf{SeekJudge} framework, whose four agents, a Condense, a Ground, a Seek and an Analyze agent, reach a verdict through a Seek--Analyze loop over the trajectory.
To train a specialized model on densely labeled trajectories, we propose a seed-driven distillation pipeline that expands a few human-labeled seed trajectories into $10$K such trajectories.
To evaluate step-level judgments, we build CUAStepBench, a human-annotated benchmark that pairs trajectory verdicts with dense step labels.
Beyond accuracy, SeekJudge costs far less than a closed-source large model. We further propose rollout overlap, a reward-server design that reduces the overhead of a reward model in RL training.
To our knowledge, SeekJudge is the first model-based reward to match or surpass native rule-based reward in online RL, measured by downstream success rate on held-out RL test goals.
In offline judging, SeekJudge-9B also exceeds rule-based evaluation by $9.5$ F1 on AgentRewardBench.

%% file: src/introduction.tex
\section{Introduction}
\label{sec:intro}

Computer-use agents (CUAs) have moved from constrained synthetic environments such as MiniWoB \citep{shi2017wob} to open-ended tasks across web, mobile and desktop platforms \citep{zhou2023webarena,deng2023mind2web,xie2024osworld,rawles2024androidworld}, and their trajectories have grown from tens of steps to hundreds of steps \citep{yuan2026osworld2}. Deciding whether the long trajectory an actor produces fulfills the instruction determines how we measure a model and how well downstream reinforcement learning or rejection fine-tuning works.

For a long time, judging a trajectory has relied mainly on rule-based evaluation \citep{xie2024osworld,rawles2024androidworld,zhou2023webarena,wang2026cuagymscalingverifiabletraining}. Rule-based evaluation suffers from three problems.
\textbf{(1) A rule disagrees with human intention.} Many valid answers can satisfy the same goal, yet a rule cannot enumerate them all \citep{abc2025}. The extreme case is an open-ended instruction such as drawing a picture, where acceptable outputs are unbounded and no rule can judge the result. On an existing CUA benchmark, a blind human re-check of every trajectory where rule-based evaluation and a Codex judge give opposite verdicts agrees with the judge more often than with the rule, as Figure~\ref{fig:multiimg}(a) shows.
\textbf{(2) A rule needs a dedicated parser per application and breaks across application versions.} A rule reads internal state through layout or internal APIs, so each new application needs its own parser or a mock app, and the parser stops working once the application moves to a new version.
\textbf{(3) A rule's ground-truth answer becomes outdated.} Even when the application stays the same, the content of a live website keeps changing. A rule relies on a predefined answer, so once that content changes the answer drifts or the task becomes unsolvable, for example when a task asks for the CEO of a company and the company later appoints a new one. Covering live online services therefore incurs a heavy maintenance cost \citep{xue2025an}.

These problems of rules motivate a shift from rule-based to model-based evaluation, but existing frameworks have not yet made this paradigm mainstream, because their judging accuracy remains limited \citep{cuajudge2025,osthemis2025,cuaverifierbench2026}. These frameworks put many images into one context while only a small high-fidelity fragment determines the verdict, and thus they may overlook the decisive evidence. Even though the decisive image is always present, performance degrades increasingly with the number of trajectory images added to the same forward pass, as Figure~\ref{fig:multiimg}(b) shows.

\input{src/figures/fig_multiimg}

We decompose judging a trajectory into two subtasks: localization, which identifies the image holding the decisive evidence, and extraction, which reads the decisive detail from that image accurately. Based on this decomposition, we propose \textbf{SeekJudge}, a multi-agent model-based judge that separates the two subtasks. Its Seek agent localizes on a text timeline, and its Analyze agent extracts the decisive detail from one image at a time. Its Condense agent records the state transition of each step and its Ground agent identifies the element that each action hits, which together build this timeline from the trajectory. SeekJudge also has the fine-grained scoring, low cost and low latency that a reward model inside RL needs:

\begin{itemize}[leftmargin=1.5em]
  \item \textbf{On judging quality}, SeekJudge returns a continuous score for the trajectory and classifies every step, which gives a graded reward and step-level judgments.
  \item \textbf{On efficiency}, the whole system is a single $9$B open-source model, hundreds of times cheaper than a closed-source large model, and its peak context stays $4$--$6\times$ smaller than that of existing model-based judges. Rollout overlap further runs part of the judging while the environment executes actions, so the reward model adds less time to the rollout.
\end{itemize}

In summary, our contributions are as follows. \textbf{(1)} We propose SeekJudge, a multi-agent framework that separates localization from extraction, together with rollout overlap that makes it faster inside reinforcement learning. \textbf{(2)} We construct CUAStepBench, a CUA reward benchmark that pairs human trajectory verdicts with dense step labels, spanning $177$ applications across platforms. \textbf{(3)} We train SeekJudge-9B for the framework. To our knowledge, it is the first model-based reward to match or surpass native rule-based reward on downstream test success in online RL. It also exceeds existing model-based judges on offline reward benchmarks.

%% file: src/figures/fig_multiimg.tex
\begin{figure}[t]
  \centering
  \includegraphics[width=1.0\linewidth]{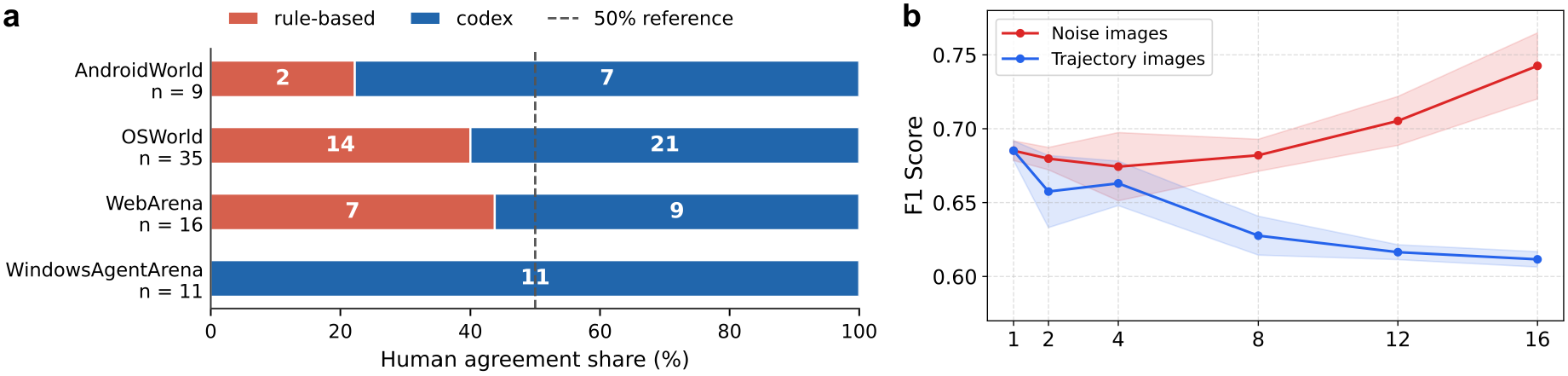}
  \caption{\textbf{(a)} A Codex judge runs over about $1$K trajectories of OmniGUIRewardBench. On the $71$ trajectories where it and rule-based evaluation disagree, a blind human re-check agrees with the judge (\emph{blue}) more often than with the rule (\emph{orange}). \textbf{(b)} With the decisive screenshot always present, padding with other trajectory images (\emph{blue}) lowers F1 while equal-token noise images (\emph{red}) do not. Details are in Appendix~\ref{app:multiimg}.}
  \label{fig:multiimg}
\end{figure}

%% file: src/related_work.tex
\section{Related Work}

\paragraph{From rule-based to model-based reward.} Deciding whether a trajectory fulfills its instruction underlies both benchmark scores and the RL reward. Both mainly rely on rule-based evaluation \citep{xie2024osworld,rawles2024androidworld,zhou2023webarena,wang2026cuagymscalingverifiabletraining}. These rules can themselves be wrong, and OSWorld has fixed its evaluators after users reported errors \citep{osworldverified2025}. Model-based judges replace the rule with a model. CUAJudge \citep{cuajudge2025} judges whole trajectories in multi-image forward passes, where the decisive screenshot competes with many others, OS-Themis \citep{osthemis2025} runs a multi-agent critic whose context grows with trajectory length, and ProRe \citep{dai2026prore} and VAGEN \citep{cui2026vagen} probe the live environment after the episode ends. As RL rewards, OpenWebRL-Judge \citep{openwebrl2026}, OS-Themis and VAGEN are compared only with other model-based rewards or the untrained policy, WebJudge \citep{xue2025an} and ProRe leave RL to future work, and StainFlow \citep{hao2026stainflow} and ADMIRE \citep{zheng2026admire} add their rewards to the rule-based reward of AndroidWorld rather than replacing it. PRO-CUA \citep{he2026procua} outperforms a rule-based baseline in web RL, but that rule matches reference actions and a GPT-5 judge scores success. To our knowledge, this paper is the first to train with a model-based reward in place of the native rule-based reward and compare the two.

\paragraph{Step-level reward models.} A trajectory verdict does not say which steps went wrong. Process reward models score each step in mathematics \citep{lightman2023verify,wang2024mathshepherd} and web navigation \citep{chae2025webshepherd}. For CUAs, the World-State-Model judge of SEAgent \citep{sun2025seagent} covers only narrow Chrome tasks, GUI-Owl \citep{ye2025guiowl} embeds step critics inside the actor pipeline, and OpenCUA \citep{wang2025opencua} sees only the context up to each step, while VLM rewards for GUI RL \citep{yang2025zerogui,qi2025webrl,bai2024digirl,sumyk2026rlcuaautonomouseval} and trajectory filtering \citep{pan2024autonomous,he2024webvoyager} are generic judges \citep{zheng2023judging} without step labels. SeekJudge classifies every step from a timeline of the whole trajectory.

\paragraph{Reward benchmarks.} Evaluating step judgments needs step labels, which AgentRewardBench \citep{lu2025agentrewardbench} and OmniGUIRewardBench \citep{osthemis2025} lack. CUARewardBench \citep{cuarewardbench2025} labels sparse key actions on Ubuntu only, and OS-Critic Bench \citep{wu2026osoracle} labels unexecuted candidate actions without a trajectory verdict. CUAStepBench labels every executed step together with the trajectory verdict.

%% file: src/method.tex
\section{Method}
\label{sec:method}

\subsection{The SeekJudge Framework}
\label{sec:framework}

SeekJudge realizes localization and extraction as a multi-agent framework whose four agents share one backbone model (Figure~\ref{fig:framework}). The Condense and Ground agents first compress the whole trajectory into a compact text timeline. The Seek agent takes this timeline as its initial context and runs a Seek--Analyze loop, querying the Analyze agent for the evidence that it needs before giving \mbox{its judgment}.

\input{src/figures/fig_framework}

\paragraph{Seek agent.} The Seek agent controls the loop and is the only agent that keeps state across the rounds of one judgment. When the condensed timeline holds enough evidence for a verdict, it outputs the seven dimension scores of the trajectory and a step class for each step. When the timeline lacks that evidence, it selects one screenshot, sends a focused question about it to the Analyze agent, and appends the answer to its context. The loop runs for at most $8$ rounds.

\paragraph{Analyze agent.} The Analyze agent answers one focused question about the single screenshot that the Seek agent selects and returns the decisive detail as text. Each answer sees only that \mbox{question and image}.

\paragraph{Condense and Ground agents.} For each step the Condense agent reads the screenshots before and after the action and records the state transition in a few lines of text as the T entry of the timeline. The Ground agent reads the screenshot before the action together with the raw action that the actor executed, such as \texttt{click 128,453}, identifies the element that those coordinates hit, and records it as the A entry of the timeline.

Separating localization from extraction gives three properties. \textbf{(1)} Every agent reads at most two images at a time, at high fidelity, with one screenshot for Analyze and Ground and two for Condense. Appendix~\ref{app:multiimg} shows that, even with the decisive screenshot always in the input, a verdict degrades as more screenshots of the same trajectory share one forward pass. \textbf{(2)} Evaluation scales to longer trajectories because Condense and Ground compress every step into a few lines of text, and Section~\ref{sec:cost} measures a peak context $4$--$6\times$ smaller than that of existing judges. \textbf{(3)} Compute adapts to length because the Seek--Analyze loop issues only the tool calls to Analyze that a case needs, and Appendix~\ref{app:adaptive} counts these tool calls against trajectory length.

\subsection{Dense Supervision via Seed-Driven Distillation}
\label{sec:datagen}

To train the single $9$B backbone that the four agents share, we build supervised data for each of the four roles through a seed-driven construction with a Seed and a Train Stage, and a Judge Stage reduces the dense outputs to scalars, as Figure~\ref{fig:pipeline} summarizes.

\input{src/figures/fig_pipeline}

\paragraph{Dense label design.} A single true/false label is coarse in two ways. It cannot rank several trajectories of one task beyond pass or fail, and it merges the distinct axes of success or failure, such as the fraction of the task completed or the confidence of the judgment, into one bit. To separate these axes, we score each trajectory on seven dimensions, namely goal infeasibility, goal ambiguity, exploration degree, path redundancy, positive evidence reliability, negative evidence reliability, and completion progress, each on a $0$--$100$ scale. We assign each step one of nine classes instead of a true/false mark.

\paragraph{Seed Stage.}
This stage tunes the two prompts that DeepSeek-V3.2 uses as the Seek agent. The criteria prompt defines what counts as success and is searched against about $20$ trajectories that a human labels with seven dimension scores and step classes. The seek prompt specifies what to ask the Analyze agent and when to stop and is searched against the progress descriptions that Codex with GPT-5.5 writes for about $1$K trajectories. Both searches run in Claude Code, and Appendix~\ref{app:leakage:prompt} describes how neither search reads benchmark ground truth.

\paragraph{Train Stage.} We run the full SeekJudge pipeline with the Seed Stage prompts, DeepSeek-V3.2 as the Seek agent and Gemini-3.0-Flash as the Analyze, Condense and Ground agents, and record the input and output of every agent step as distillation data. Supervised fine-tuning (SFT) of Qwen3.5-9B on this data gives SeekJudge-9B. Appendix~\ref{app:sft} lists the SFT data and hyperparameters.

\paragraph{Judge Stage.} The framework outputs the seven dimension scores of the trajectory and a step class for each step, and the Judge Stage reduces these scores and classes to the scalars that offline evaluation and online RL use.
A small gradient-boosted regressor maps the seven dimension scores to a trajectory score $S^{\mathrm{traj}}$, and each of the nine step classes maps to a step score $S^{\mathrm{step}}$ set by a human rubric. Offline evaluation and RL each use one fit of the regressor, and Appendix~\ref{app:leakage:fit} lists both fits and probes the offline fit for leakage. With the offline fit alone in place of the RL fit, SeekJudge still surpasses rule-based reward in RL on UI-TARS-1.5-7B Impress, as Appendix~\ref{app:rlfit} shows.
\begin{itemize}[leftmargin=1.5em]
  \item \textbf{For offline evaluation}, one threshold on $S^{\mathrm{traj}}$, shared by every benchmark and backbone, gives the trajectory verdict, and the three harmful classes count as erroneous steps while the remaining six count as correct. Appendix~\ref{app:stepclass} defines the nine classes, marks which count as correct, and lists their step scores.
  \item \textbf{For online RL}, the reward server returns, for rollout $i$ of a group of $G$ rollouts that share one goal, the trajectory score $S^{\mathrm{traj}}_i$ and the step scores $S^{\mathrm{step}}_{i,1},\dots,S^{\mathrm{step}}_{i,N_i}$ of its $N_i$ steps. The trajectory reward adds the mean step score to the trajectory score with weight $w_{\mathrm{step}}=2$,
  \begin{equation}
  R_i = S^{\mathrm{traj}}_i + \frac{w_{\mathrm{step}}}{N_i}\sum_{k=1}^{N_i} S^{\mathrm{step}}_{i,k}.
  \label{eq:reward}
  \end{equation}
  GRPO normalizes $R_i$ within the group and assigns the resulting advantage to every step of rollout $i$. This advantage carries no step index $k$, and the step scores enter it only through their mean in Eq.~\ref{eq:reward}. This is the weak form of a process reward, with the step term assisting the trajectory-level reward, and it is the default throughout the paper. Appendix~\ref{app:reward} gives the implementation, and Appendix~\ref{app:stepreward} tests the strong form, which injects the step scores into the advantage of every step.
\end{itemize}

\paragraph{Training data.} We collect training trajectories for $10{,}905$ distinct tasks across web, desktop (Windows, Ubuntu and macOS) and Android. Every task goal is a human-written description, and the set has zero overlap with any reward benchmark that we tested or any OSWorld goal. Appendix~\ref{app:data} gives the composition of the set by environment and application.

\subsection{Rollout-Overlapped, Zero-Client-State Reward Server}
\label{sec:infra}

\input{src/figures/fig_server}

Serving a reward model inside RL raises a scheduling problem. In a GUI rollout, the environment executes each action of the multi-turn loop while the actor's GPUs idle, and a judge that runs only after the trajectory ends adds its whole latency to every rollout. We design the reward server around this loop with two mechanisms. Rollout overlap fills the idle window with judging, and zero client state keeps the state management that this overlap needs out of the RL-side code.

\paragraph{Rollout overlap.} We overlap judging with the rollout so the reward model adds less latency to training. Operations that do not depend on later actions, such as Condense and Ground here or key-information identification in CUAJudge, run as soon as their screenshots and action arrive, in the idle GPU windows while the environment executes the following actions (Figure~\ref{fig:server}).

\paragraph{Zero client state.} We move the scheduling complexity of rollout overlap from the RL side to the reward server. The RL side keeps no reward-related state and streams each screenshot and action as soon as it is produced, in any order within one rollout or across rollouts, and the server reorders the stream and schedules the overlapped computation. The reward model thus attaches to an existing RL framework without changes to the RL-side code.

%% file: src/figures/fig_framework.tex
\begin{figure}[t]
  \centering
  \includegraphics[width=0.92\linewidth]{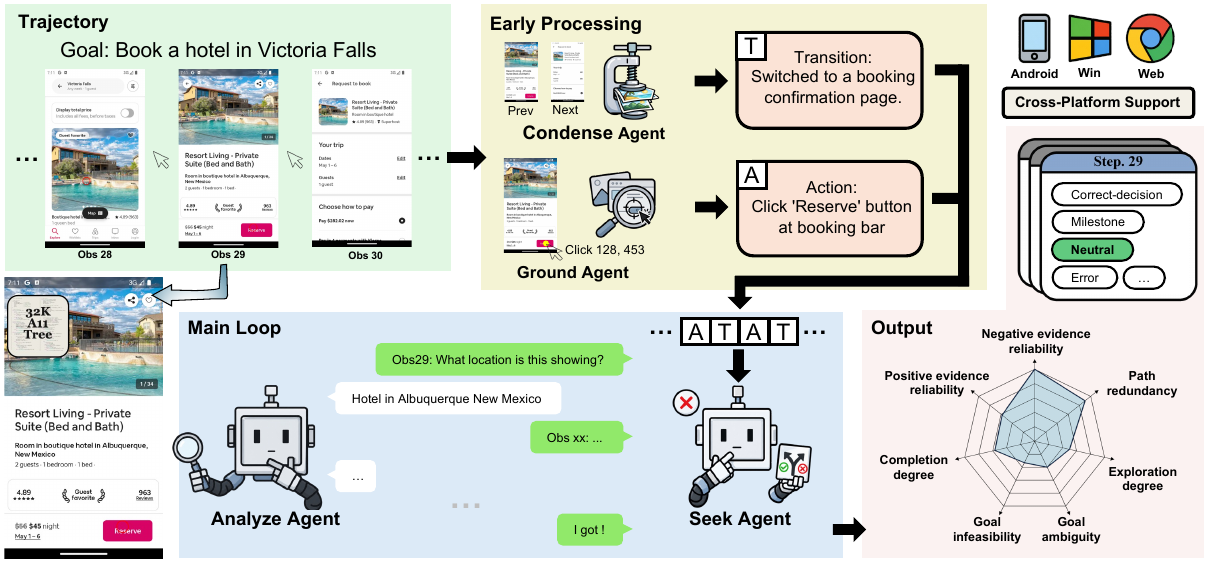}
  \captionsetup{singlelinecheck=false,justification=raggedright}
  \caption{The SeekJudge framework. Four agents share one backbone. Condense and Ground write a text timeline with one transition entry (T) and one grounded-action entry (A) per step.}
  \label{fig:framework}
\end{figure}

%% file: src/figures/fig_pipeline.tex
\begin{figure}[t]
  \centering
  \includegraphics[width=0.88\linewidth]{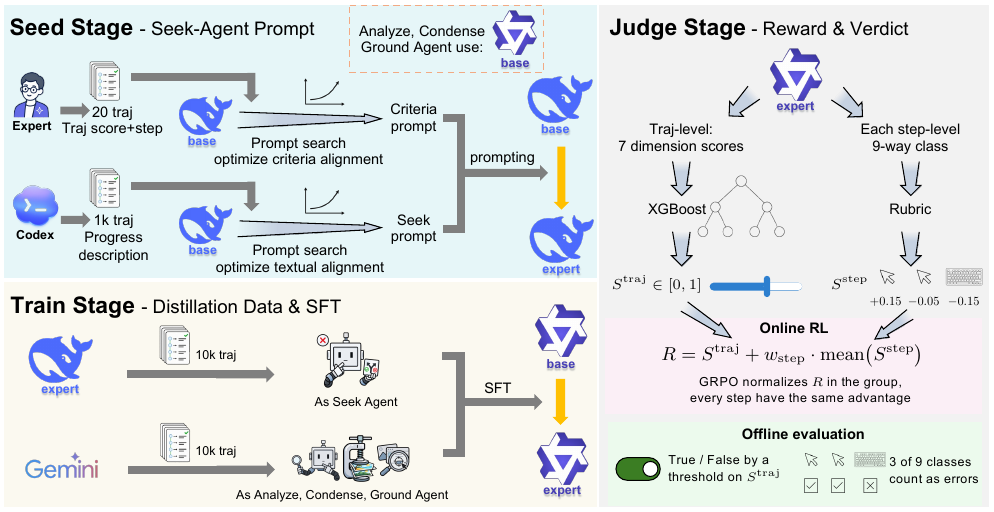}
  \caption{The seed-driven construction of training data, the evaluation process, and the reduction of the dense labels to a trajectory verdict and step labels for offline evaluation and to one reward per rollout for online RL (Eq.~\ref{eq:reward}).}
  \label{fig:pipeline}
\end{figure}

%% file: src/figures/fig_server.tex
\begin{figure}[t]
  \centering
  \includegraphics[width=0.9\linewidth]{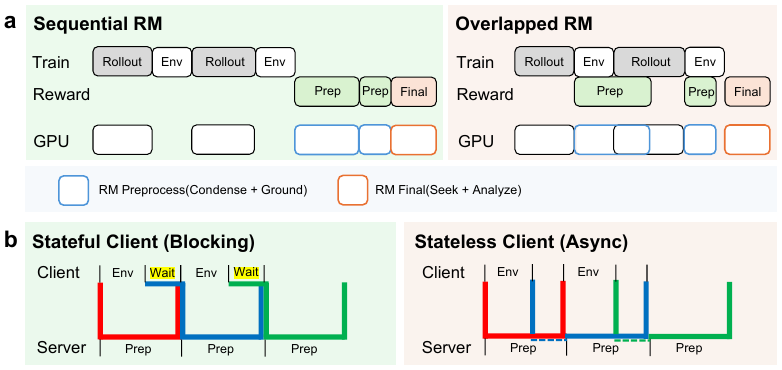}
  \caption{Reward server. \textbf{(a)} A sequential reward model judges only after the trajectory ends, while SeekJudge runs Condense and Ground during the rollout, leaving only the final stage on the critical path. \textbf{(b)} A stateful client blocks between steps to keep requests ordered, while a zero-state client streams out of order and the server reorders.}
  \label{fig:server}
\end{figure}

%% file: src/experiments.tex
\section{Experiments}
\label{sec:experiments}

\subsection{CUAStepBench} %
\label{sec:bench}

\input{src/tables/tab_bench_compare}

To evaluate the step-level judgments of a reward model together with its trajectory verdicts, we build CUAStepBench, a human-annotated benchmark of $278$ tasks over $177$ applications. Every trajectory carries a human verdict, and beyond that verdict each step receives a human binary label of whether it is correct. Table~\ref{tab:bench_compare} places CUAStepBench among existing reward benchmarks. Every task goal in CUAStepBench comes from a human-written task description.

We keep hard cases so that the benchmark separates judges. Two of the seven dimension scores, negative and positive evidence reliability, are the confidence that the trajectory failed and the confidence that it succeeded. We run SeekJudge with its teacher models over the candidate pool and keep the tasks with both confidences low. Low confidence on both sides marks a trajectory whose outcome the screenshots do not clearly determine. We then balance the retained tasks across trajectory lengths, applications and task types. Appendix~\ref{app:bench} describes this curation from the candidate pool to the final $278$ tasks. Only about $500$ of the roughly $5$K step labels are error steps, so we report F1 rather than accuracy at the step level, and F1 at the trajectory level as well for consistency.

\subsection{Training and Evaluation Details}
\label{sec:training}

\input{src/tables/tab_rl_main}

\input{src/tables/tab_offline_cuastepbench}

\input{src/tables/tab_offline_traj}

All RL environments come from the Chrome, Impress and OS domains of OSWorld \citep{xie2024osworld}, with $12$ to $24$ training and test tasks per domain. We train a separate policy for every domain with GRPO, and every reward is compared under identical training within one domain. Appendix~\ref{app:training} lists the hyperparameters and the evaluation protocol. We report test success as the mean over the test evaluations taken every $15$ steps during training and training success as the rule-measured rate on the training tasks averaged over training steps. Across three pairs of repeated runs, the pooled standard deviation is about $2.0$ points for test success and $2.2$ for training success, so a reward gap averaged over two backbones and three domains has a standard error of $1.15$ points.

\subsection{Reinforcement Learning Main Results}
\label{sec:rl_main}

Table~\ref{tab:rl_main} reports the test success rate of policies trained under each reward model. The results show that \textbf{(1)} Averaged over the six backbone-domain settings, SeekJudge exceeds native rule-based reward by $3.3$ points of test success (about $2.9$ times the standard error), while rule-based reward reaches the highest training success rate because the Train column is its own training signal. \textbf{(2)} Among model-based rewards, SeekJudge gives the highest average test success on both backbones. CUAJudge and OS-Themis enter RL through their native binary verdicts, and SeekJudge leads them by combining a finer-grained reward with a stronger judge. Appendix~\ref{app:reward_hacking} measures reward hacking against the rule-based evaluation during training.

\subsection{Offline Reward Benchmark Evaluation}
\label{sec:offline}

\input{src/figures/fig_cost}

We evaluate SeekJudge as a static judge on three reward benchmarks against CUAJudge, OS-Themis and the OS-Shepherd-9B judge from OSReward \citep{sun2026osrewardinstitutingstandardizedevaluation}, with the large-model CUAJudge on GPT-5-mini, the model of its release. CUAStepBench scores both the trajectory verdict and the step labels, while AgentRewardBench and OmniGUIRewardBench score the trajectory verdict alone. The results show that \textbf{(1)} SeekJudge-9B is competitive with every baseline except the agentic Codex run. It exceeds the Rule baseline on AgentRewardBench by $9.5$ F1, reaches a higher trajectory F1 than the GPT-5-mini-based CUAJudge on all three benchmarks, and matches its step-level F1 on CUAStepBench, $38.0$ against $37.9$.
\textbf{(2)} With the untrained Qwen3-VL-8B backbone, the SeekJudge framework already exceeds OS-Themis on the same backbone in trajectory F1 on all three benchmarks, with high recall at the cost of precision.
All SeekJudge entries use one shared threshold, and their precision and recall depend on it. Appendix~\ref{app:roc} plots the threshold-free ROC curve of SeekJudge-9B against every baseline's operating point.

\subsection{Efficiency: Cost, Context and Training Overhead}
\label{sec:cost}

\input{src/tables/tab_overlap_speed}

\textbf{Cost.} We compute the cost of every judge on the same $43$ cases under one cost model that converts open-source and closed-source calls to a common dollar cost per token, with the token prices and the cache rule given in Appendix~\ref{app:cost}. Figure~\ref{fig:cost} shows that SeekJudge judges at the lowest cost at every image count, and its cost gap to every other judge widens as the number of images grows. CUAJudge incurs extra GPT-5-mini calls, and the agentic Codex baseline costs two orders of magnitude more.

\textbf{Context.} The peak context of a judge sets the longest trajectory that fits in its context window. Figure~\ref{fig:max_context} plots, for each case, the token count of the largest single request that a judge issues against the number of images. At $52$ images the fitted lines give about $12$K tokens for SeekJudge, $48$K for OS-Themis and $80$K for CUAJudge, about $4\times$ and $6\times$ that of SeekJudge. Extrapolated, the lines reach $128$K tokens at about $90$ images for CUAJudge, $160$ for OS-Themis and $700$ for SeekJudge.

\textbf{Training overhead.} We train Qwen3-VL-8B with the SeekJudge reward in each domain twice, judging only after each trajectory ends or with the rollout overlap of Section~\ref{sec:infra}, and measure the round time and the overhead outside the actor's and environment's own execution. Most of this overhead is the time that training waits for the reward model to return its judgments. Table~\ref{tab:overlap_speed} shows that rollout overlap reduces the overhead by $56.8$\% overall, from $18.3$~s to $7.9$~s per round.

\subsection{Ablations}
\label{sec:ablation}

\begin{table}[t]
  \captionsetup{justification=raggedright,singlelinecheck=false}
  \begin{minipage}[t]{0.48\linewidth}
    \input{src/tables/tab_reward_granularity}
  \end{minipage}
  \hfill
  \begin{minipage}[t]{0.48\linewidth}
    \input{src/tables/tab_framework_ablation}
  \end{minipage}
\end{table}

\paragraph{Reward granularity.} Table~\ref{tab:reward_granularity} trains UI-TARS-1.5-7B on Impress with a Boolean, a continuous, and a continuous plus step-level SeekJudge reward. Adding the step term of Eq.~\ref{eq:reward} to the continuous score gives the default reward and raises test success from $33.62$ to $38.84$, the highest in the table.

\paragraph{Framework ablation.} Table~\ref{tab:framework_ablation} removes components of SeekJudge at inference and evaluates each variant on CUAStepBench with a pair of strong models. We use this pair rather than SeekJudge-9B to avoid a distribution shift caused by evaluating SeekJudge-9B differently from how it was trained. Condense and the Seek--Analyze loop both contribute. Without the loop, removing Condense lowers trajectory F1 by $5.5$ points and step F1 by $4.4$ points. Adding the loop to Condense and Ground raises trajectory F1 by $2.3$ points and step F1 by $7.0$ points.

%% file: src/tables/tab_bench_compare.tex
\begin{table}[t]
  \centering
  \caption{CUAStepBench compared with existing reward benchmarks. \emph{Traj. labels} gives the source of the trajectory verdict, and \emph{Step labels} gives which steps carry a label and how many.}
  \label{tab:bench_compare}
  \small
  \setlength{\tabcolsep}{5pt}
  \begin{tabular}{lccc}
    \toprule
    Benchmark & Traj. labels & Step labels & Labeled steps \\
    \midrule
    AgentRewardBench \citep{lu2025agentrewardbench} & Human & None & -- \\
    OmniGUIRewardBench \citep{osthemis2025} & Rule & None & -- \\
    CUARewardBench \citep{cuarewardbench2025} & Human & Key actions only & $346$ \\
    CUAVerifierBench \citep{cuaverifierbench2026} & Human & Whole-trajectory process only & -- \\
    OS-Critic Bench \citep{wu2026osoracle} & None & Unexecuted candidate actions & $738$ \\
    \textbf{CUAStepBench (ours)} & Human & Every executed step & $5{,}634$ \\
    \bottomrule
  \end{tabular}
\end{table}

%% file: src/tables/tab_rl_main.tex
\begin{table}[t]
  \centering
  \caption{Reinforcement-learning main results. Training (Train) and test (Test) success rates in \%, both measured by rule-based evaluation under the protocol of Section~\ref{sec:training}.}
  \label{tab:rl_main}
  \setlength{\tabcolsep}{4pt}
  \begin{tabular}{llcccccc!{\hspace{6pt}}cc}
    \toprule
    & & \multicolumn{2}{c}{Chrome} & \multicolumn{2}{c}{Impress} & \multicolumn{2}{c!{\hspace{6pt}}}{OS} & \multicolumn{2}{c}{Avg} \\
    \cmidrule(lr){3-4} \cmidrule(lr){5-6} \cmidrule(l{5pt}r{11pt}){7-8} \cmidrule(lr){9-10}
    Backbone & Reward & Train & Test & Train & Test & Train & Test & Train & Test \\
    \midrule
    \multirow{4}{*}{UI-TARS-1.5-7B}
             & Rule-based   & \textbf{41.33} & 12.75 & \textbf{43.00} & 30.43 & \textbf{55.32} & 25.56 & \textbf{46.55} & 22.91 \\
             & CUAJudge     & 34.15 & \underline{13.91} & 29.18 & \underline{33.48} & \underline{49.83} & \underline{26.11} & 37.72 & \underline{24.50} \\
             & OS-Themis     & 12.64 & 10.14 & 19.59 & 32.46 & 20.45 & 25.00 & 17.56 & 22.53 \\
             & SeekJudge      & \underline{37.40} & \textbf{16.23} & \underline{30.06} & \textbf{38.84} & 49.11 & \textbf{28.89} & \underline{38.86} & \textbf{27.99} \\
    \midrule
    \multirow{4}{*}{Qwen3-VL-8B}
             & Rule-based   & \textbf{54.89} & \underline{14.49} & \textbf{43.84} & \underline{49.28} & \textbf{69.73} & \textbf{26.67} & \textbf{56.15} & \underline{30.15} \\
             & CUAJudge     & 47.56 & 14.20 & 27.35 & 40.58 & 61.57 & 20.56 & 45.49 & 25.11 \\
             & OS-Themis     & 42.57 & 11.01 & 20.58 & 40.87 & 65.14 & \underline{25.00} & 42.76 & 25.63 \\
             & SeekJudge      & \underline{48.43} & \textbf{20.58} & \underline{32.19} & \textbf{50.14} & \underline{66.11} & 24.17 & \underline{48.91} & \textbf{31.63} \\
    \bottomrule
  \end{tabular}
\end{table}

%% file: src/tables/tab_offline_cuastepbench.tex
\begin{table}[t]
  \centering
  \caption{Offline evaluation on CUAStepBench, trajectory- and step-level Precision, Recall and F1 in \%, with step-level metrics over error steps. All judge rows are our runs, and -- marks judges that give no step labels. $^{\dagger}$ marks frameworks without native step-level judgments, whose step labels come from the step prompt of SeekJudge (Appendix~\ref{app:context}) on the same base model. DS + Gemini is DeepSeek-V4-Pro-0423 as the Seek agent and Gemini-3.0-Flash as the Analyze agent.}
  \label{tab:offline_cuastepbench}
  \begin{tabular}{ll ccc ccc}
    \toprule
    & & \multicolumn{3}{c}{Trajectory-level} & \multicolumn{3}{c}{Step-level} \\
    \cmidrule(lr){3-5} \cmidrule(lr){6-8}
    Framework & Model & Prec & Recall & F1 & Prec & Recall & F1 \\
    \midrule
    Human expert         & --             & 80.3 & 84.1 & 82.2 & -- & -- & -- \\
    \specialrule{\lightrulewidth}{\aboverulesep}{0pt}
    \multicolumn{8}{l}{\textit{Large models}} \\
    Codex                & GPT-5.5      & 87.2 & 70.3 & 77.9 & -- & -- & -- \\
    CUAJudge$^{\dagger}$             & GPT-5-mini    & 78.6 & 60.7 & 68.5 & 49.5 & 30.6 & 37.9 \\
    SeekJudge            & DS + Gemini    & 64.0 & 90.3 & 74.9 & 48.7 & 43.0 & 45.7 \\
    \midrule
    \multicolumn{8}{l}{\textit{Small models}} \\
    OSReward             & OS-Shepherd-9B & 74.2 & 63.4 & 68.4 & -- & -- & -- \\
    CUAJudge$^{\dagger}$             & Qwen3-VL-8B    & 72.0 & 62.1 & 66.7 & 66.7 & 12.6 & 21.2 \\
    OS-Themis$^{\dagger}$             & Qwen3-VL-8B      & 71.3 & 51.5 & 59.8 & 17.0 & 14.2 & 15.5 \\
    SeekJudge                       & Qwen3-VL-8B   & 56.6 & 96.4 & \underline{71.3} & 41.4 & 20.4 & \underline{27.3} \\
    SeekJudge                       & SeekJudge-9B  & 68.6 & 80.8 & \textbf{74.2} & 44.5 & 33.1 & \textbf{38.0} \\
    \bottomrule
  \end{tabular}
\end{table}

%% file: src/tables/tab_offline_traj.tex
\begin{table}[t]
  \centering
  \caption{Offline evaluation on AgentRewardBench and OmniGUIRewardBench, trajectory-level results in \%. $^{\dagger}$ marks results and model names quoted from their source papers. Rule reproduces the execution-rule labels of OmniGUIRewardBench by construction, so its entries are italic and excluded from the marking.}
  \label{tab:offline_traj}
  \begin{tabular}{ll ccc ccc}
    \toprule
    & & \multicolumn{3}{c}{AgentRewardBench} & \multicolumn{3}{c}{OmniGUIRewardBench} \\
    \cmidrule(lr){3-5} \cmidrule(lr){6-8}
    Framework & Model & Prec & Recall & F1 & Prec & Recall & F1 \\
    \midrule
    Rule$^{\dagger}$             & --            & 83.8 & 55.9 & 67.1 & \textit{100.0} & \textit{100.0} & \textit{100.0} \\
    WebJudge$^{\dagger}$         & o4-mini       & 82.0 & 47.8 & 60.4 & -- & -- & -- \\
    WebJudge$^{\dagger}$         & Qwen + o4-mini  & 75.7 & 58.0 & 65.6 & -- & -- & -- \\
    World-State-Model$^{\dagger}$ & WSM-7B            & 71.2 & 72.2 & \underline{71.7} & -- & -- & -- \\
    CUAJudge & GPT-5-mini    & 75.2 & 53.6 & 62.6 & 90.5 & 78.0 & \underline{83.8} \\
    OS-Themis & Qwen3-VL-8B      & 73.1 & 42.4 & 53.6 & 88.4 & 63.1 & 73.7 \\
    SeekJudge                       & Qwen3-VL-8B   & 46.3 & 90.1 & 61.2 & 67.3 & 90.8 & 77.3 \\
    SeekJudge                       & SeekJudge-9B  & 74.2 & 79.2 & \textbf{76.6} & 85.3 & 85.0 & \textbf{85.1} \\
    \bottomrule
  \end{tabular}
\end{table}

%% file: src/figures/fig_cost.tex
\begin{figure}[t]
  \centering
  \begin{minipage}[t]{0.44\linewidth}
    \centering
    \includegraphics[width=\linewidth]{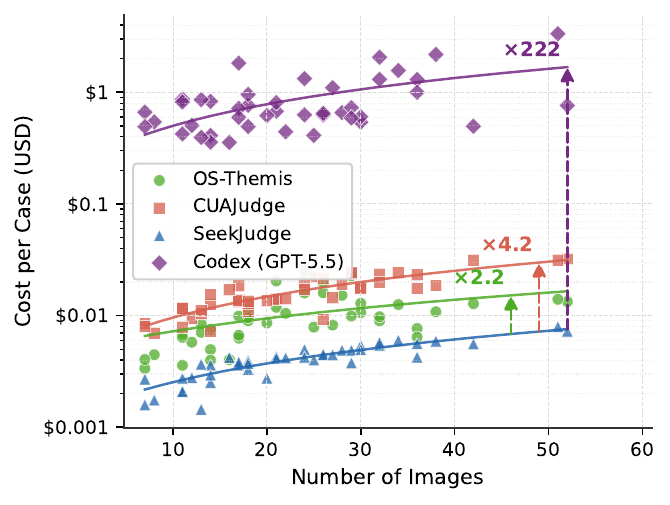}
    \caption{Per-case judging cost against the number of images, one marker per case with a linear fit per judge on a log cost axis.}
    \label{fig:cost}
  \end{minipage}
  \hfill
  \begin{minipage}[t]{0.44\linewidth}
    \centering
    \includegraphics[width=\linewidth]{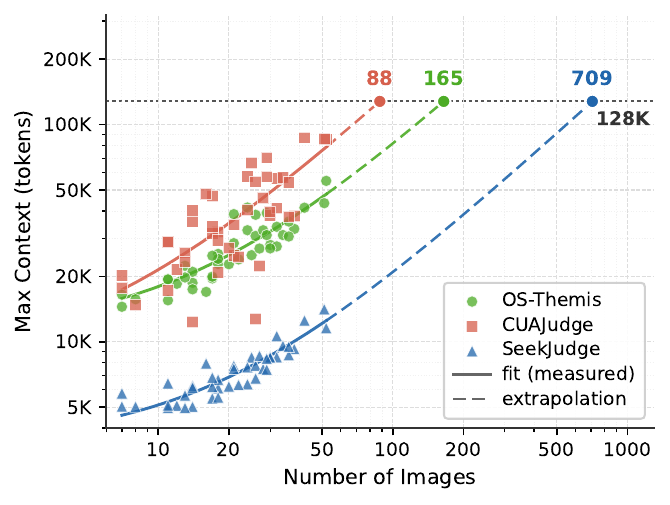}
    \caption{Peak context size against the number of images, continued dashed to the $128$K window.}
    \label{fig:max_context}
  \end{minipage}
\end{figure}

%% file: src/tables/tab_overlap_speed.tex
\begin{table}[t]
  \centering
  \caption{Rollout time per interaction round in seconds, Qwen3-VL-8B trained with the SeekJudge reward, judging after each trajectory ends (Sequential) or with rollout overlap. Overhead is the part of Total outside the actor's and environment's own execution. Overall averages the domains.}
  \label{tab:overlap_speed}
  \newcommand{\rel}[1]{{\small\itshape$-$#1\%}}
  \setlength{\tabcolsep}{3pt}
  \begin{tabular}{lcccccccc}
    \toprule
    & \multicolumn{2}{c}{Chrome} & \multicolumn{2}{c}{Impress} & \multicolumn{2}{c}{OS} & \multicolumn{2}{c}{Overall} \\
    \cmidrule(lr){2-3} \cmidrule(lr){4-5} \cmidrule(lr){6-7} \cmidrule(lr){8-9}
    Scheduling & Total & Overhead & Total & Overhead & Total & Overhead & Total & Overhead \\
    \midrule
    Sequential      & 61.0 & 24.1 & 55.3 & 20.6 & 59.2 & 10.2 & 58.5 & 18.3 \\
    Rollout overlap & \textbf{46.6} & \textbf{8.1} & \textbf{48.0} & \textbf{9.4} & \textbf{51.8} & \textbf{6.2} & \textbf{48.8} & \textbf{7.9} \\
    \small\itshape Reduction & \rel{23.6} & \rel{66.4} & \rel{13.2} & \rel{54.4} & \rel{12.5} & \rel{39.2} & \rel{16.6} & \rel{56.8} \\
    \bottomrule
  \end{tabular}
\end{table}

%% file: src/tables/tab_reward_granularity.tex
\centering
\caption{Reward-granularity ablation, training and test success rates in \% of \mbox{UI-TARS-1.5-7B} on the Impress domain.}
\label{tab:reward_granularity}
\setlength{\tabcolsep}{4pt}
\begin{tabular}{llcc}
  \toprule
  Reward & Granularity & Train & Test \\
  \midrule
  Rule-based & Boolean                  & \textbf{43.00} & 30.43 \\
  SeekJudge  & Boolean                  & 21.08 & 35.94 \\
  SeekJudge  & Continuous               & 24.25 & 33.62 \\
  SeekJudge  & \begin{tabular}[c]{@{}l@{}}Continuous\\+ step-level\end{tabular} & 30.06 & \textbf{38.84} \\
  \bottomrule
\end{tabular}

%% file: src/tables/tab_framework_ablation.tex
\centering
\caption{Framework ablation on a pair of strong models, \mbox{DeepSeek-V4-Pro-0423} as the Seek agent and \mbox{Gemini-3.0-Flash} as the Analyze agent, evaluated on CUAStepBench.}
\label{tab:framework_ablation}
\setlength{\tabcolsep}{4pt}
\begin{tabular}{ccc cc}
  \toprule
  \multicolumn{3}{c}{Framework component} & \multicolumn{2}{c}{F1 (\%)} \\
  \cmidrule(lr){1-3} \cmidrule(lr){4-5}
  Condense & Ground & \begin{tabular}[c]{@{}c@{}}Seek--\\Analyze\end{tabular} & Traj & Step \\
  \midrule
  \cmark & \cmark & \cmark & \textbf{74.9} & \textbf{45.7} \\
  \cmark & \cmark & \xmark & 72.6 & 38.7 \\
  \xmark & \cmark & \xmark & 67.1 & 34.3 \\
  \cmark & \xmark & \xmark & 72.3 & 40.1 \\
  \bottomrule
\end{tabular}

%% file: src/conclusion.tex
\section{Conclusion}
\label{sec:conclusion}

We revisited long-trajectory judging for computer-use agents as the decomposition into localization and extraction, showing that, even with the decisive screenshot always present, adding more screenshots of the same trajectory to one forward pass degrades the reading of the decisive evidence. SeekJudge realizes this decomposition with four agents on a single $9$B backbone. To our knowledge, it is the first model-based reward to match or surpass native rule-based reward on downstream test success in online RL, at a fraction of closed-source cost with a $4$--$6\times$ smaller peak context. SeekJudge, CUAStepBench and the rollout-overlapped reward server together make model-based reward a practical substitute for rule-based reward and open a path to reinforcement learning in environments that rules cannot instrument.

%% file: src/appendix.tex
\newenvironment{promptexcerpt}{\begin{list}{}{\setlength{\leftmargin}{1em}\setlength{\rightmargin}{1em}}\item\relax\footnotesize\ttfamily\raggedright\setlength{\parskip}{0.5em}}{\end{list}}

\input{src/appendix/roc}
\input{src/appendix/stepreward}

\input{src/appendix/multiimg}
\input{src/appendix/adaptive}
\input{src/appendix/leakage}

\input{src/appendix/rlfit}
\input{src/appendix/reward_hacking}
\input{src/appendix/cost}
\input{src/appendix/data}

\input{src/appendix/bench}

\input{src/appendix/training}
\input{src/appendix/stepclass}

\input{src/appendix/prompt}

%% file: src/appendix/roc.tex
\section{Trajectory-Level ROC Curves}
\label{app:roc}

Tables~\ref{tab:offline_cuastepbench} and~\ref{tab:offline_traj} compare judges at one operating point each, and the threshold on the regressor score determines where on its curve the point of SeekJudge lies. Figure~\ref{fig:roc} removes the threshold. It sweeps the threshold on the regressor score of SeekJudge-9B under the joint fit of Appendix~\ref{app:leakage:fit} and plots the resulting ROC curve on each benchmark, with every other judge as the single point that its binary verdict defines. The false positive rate of a point follows from its precision, recall and accuracy, and for the results quoted from source papers on AgentRewardBench, which report no accuracy, from the positive rate of the test split. The area under the curve is $0.808$ on CUAStepBench, $0.935$ on AgentRewardBench and $0.920$ on OmniGUIRewardBench. The CUAStepBench entries of Table~\ref{tab:offline_cuastepbench} average three repeats while the sweep uses one run, so the operating point of that table lies near rather than on the curve.

\input{src/figures/fig_roc}

Three readings follow. First, on AgentRewardBench every baseline lies below the curve, so at the false positive rate of each baseline some threshold of SeekJudge-9B reaches a higher recall, $70$ against $54$ at the false positive rate of CUAJudge and $80$ against $72$ at that of World-State-Model. Second, on OmniGUIRewardBench the CUAJudge point lies on the curve and the OS-Themis point below it. Third, on CUAStepBench the agentic Codex run lies above the curve and CUAJudge with GPT-5-mini close to it, while OSReward, CUAJudge with Qwen3-VL-8B and OS-Themis lie below.

%% file: src/figures/fig_roc.tex
\begin{figure}[t]
  \centering
  \includegraphics[width=\linewidth]{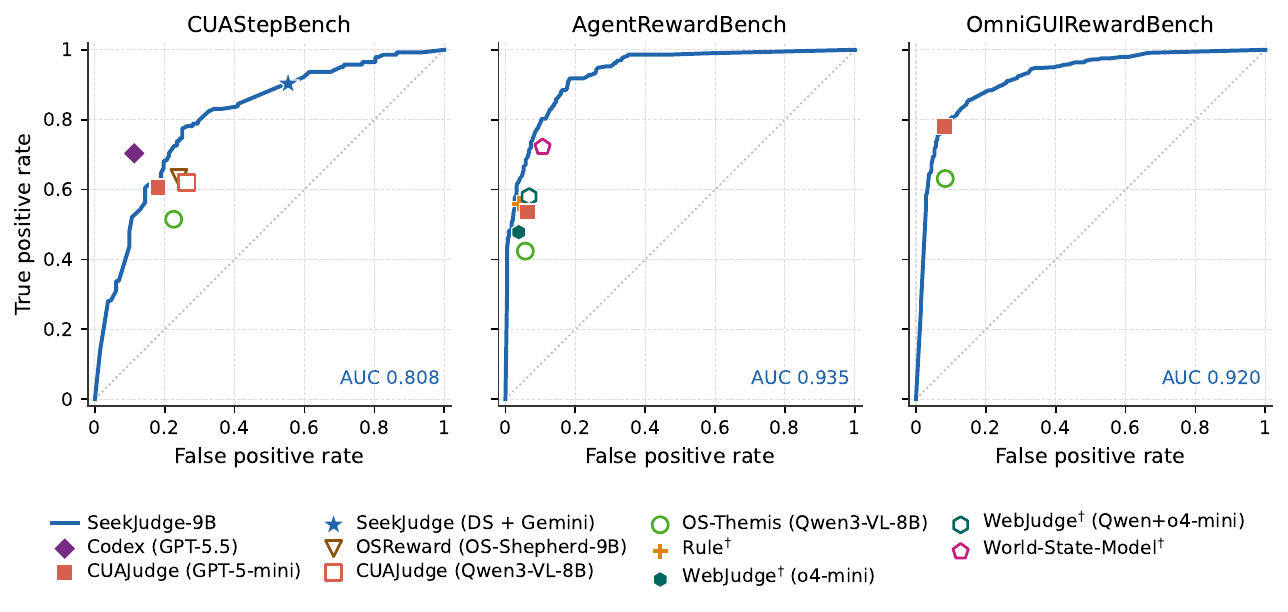}
  \caption{Trajectory-level ROC curves of SeekJudge-9B on the three offline benchmarks, from a threshold sweep over the regressor score. Every other judge returns a binary verdict and appears as one point. Filled markers are closed-source or large models and hollow markers small open models. $^{\dagger}$ marks results quoted from source papers.}
  \label{fig:roc}
\end{figure}

%% file: src/appendix/stepreward.tex
\section{Per-Step Reward Variant}
\label{app:stepreward}

SeekJudge-step keeps the step term instead of collapsing it. Every step shares the trajectory reward $R_i$ of Eq.~\ref{eq:reward} and adds its own step score, with the same weight $w_{\mathrm{step}}=2$, and the GRPO baseline pools every step of the $G=8$ rollouts in a group,
\begin{align}
R_{i,k} &= R_i + w_{\mathrm{step}}\,S^{\mathrm{step}}_{i,k}, \label{eq:stepreward}\\
\hat{A}_{i,k} &= \frac{R_{i,k} - \operatorname{mean}(\mathcal{R})}{\operatorname{std}(\mathcal{R})}, \qquad \mathcal{R} = \{R_{j,l} : 1 \le j \le G,\ 1 \le l \le N_j\}. \label{eq:stepadv}
\end{align}
Unlike Eq.~\ref{eq:adv}, the right side of Eq.~\ref{eq:stepadv} depends on $k$, and the numerator and the baseline share the per-step scale.

Table~\ref{tab:rl_step} compares the variant with SeekJudge under the protocol of Table~\ref{tab:rl_main}. The step term changes test success by $-1.7$, $+1.5$ and $-10.0$ points on UI-TARS Chrome, Impress and OS and by $-4.9$, $-4.6$ and $+10.3$ points on Qwen Chrome, Impress and OS, which lowers the UI-TARS average by $3.4$ points and leaves the Qwen average within $0.3$ point of SeekJudge. Adding the step term this way therefore gives no consistent gain. We leave per-step credit assignment to~future~work.

\input{src/tables/tab_rl_step}

%% file: src/tables/tab_rl_step.tex
\begin{table}[t]
  \centering
  \caption{Per-step reward variant (SeekJudge-step) against the trajectory reward (SeekJudge). Training (Train) and test (Test) success rates in \% under the RL protocol of Section~\ref{sec:training}.}
  \label{tab:rl_step}
  \setlength{\tabcolsep}{4pt}
  \begin{tabular}{llcccccc!{\hspace{6pt}}cc}
    \toprule
    & & \multicolumn{2}{c}{Chrome} & \multicolumn{2}{c}{Impress} & \multicolumn{2}{c!{\hspace{6pt}}}{OS} & \multicolumn{2}{c}{Avg} \\
    \cmidrule(lr){3-4} \cmidrule(lr){5-6} \cmidrule(l{5pt}r{11pt}){7-8} \cmidrule(lr){9-10}
    Backbone & Reward & Train & Test & Train & Test & Train & Test & Train & Test \\
    \midrule
    \multirow{2}{*}{UI-TARS-1.5-7B}
             & SeekJudge      & 37.40 & 16.23 & 30.06 & 38.84 & 49.11 & 28.89 & 38.86 & 27.99 \\
             & SeekJudge-step & 33.23 & 14.49 & 29.30 & 40.29 & 47.95 & 18.89 & 36.83 & 24.56 \\
    \midrule
    \multirow{2}{*}{Qwen3-VL-8B}
             & SeekJudge      & 48.43 & 20.58 & 32.19 & 50.14 & 66.11 & 24.17 & 48.91 & 31.63 \\
             & SeekJudge-step & 46.07 & 15.65 & 32.63 & 45.51 & 68.08 & 34.44 & 48.93 & 31.87 \\
    \bottomrule
  \end{tabular}
\end{table}

%% file: src/appendix/multiimg.tex
\section{Multi-Image Interference}
\label{app:multiimg}

We test whether reading a detail from an image degrades when many images share one forward pass. To isolate this extraction from localization, we keep the $123$ of $278$ CUAStepBench cases whose completion can be determined from a single screenshot, selected with the agentic Codex judge on GPT-5.5, and always place that decisive screenshot in the input. The judge under test is CUAJudge on the Qwen3-VL-8B backbone, which reads all images of a case in one forward pass. Holding the decisive screenshot fixed, we pad the input with other screenshots of the same trajectory in order (blue) or with noise images of the same token budget and no readable content (red). Figure~\ref{fig:multiimg}(b) reports F1, and Figure~\ref{fig:multiimg_pra} reports accuracy, precision and recall.

The results show that \textbf{(1)} for this judge, competing content rather than context length drives the drop. Blue and red carry the same context length, so their gap isolates trajectory content from length. Blue F1 falls from $68$ to $61$ as images accumulate, even though the decisive image is always present, while red stays level up to $8$ images and then rises, so the added trajectory detail degrades the verdict and length alone does not.
\textbf{(2)} More images bias the judge toward accepting the trajectory. Splitting blue F1 into its terms, recall rises slightly while precision drops from $56$ to $45$.

\input{src/figures/fig_multiimg_pra}

The setting matches multimodal needle-in-a-haystack retrieval, which also degrades as images accumulate \citep{wang2024mmneedle}. One possible explanation for the rise of the noise-image curve is the effect reported by \citet{jang2025expandingcomputationspacesllms}, where extra uninformative tokens widen the model's computation and add parallel scratch space.

%% file: src/figures/fig_multiimg_pra.tex
\begin{figure}[t]
  \centering
  \includegraphics[width=1.0\linewidth]{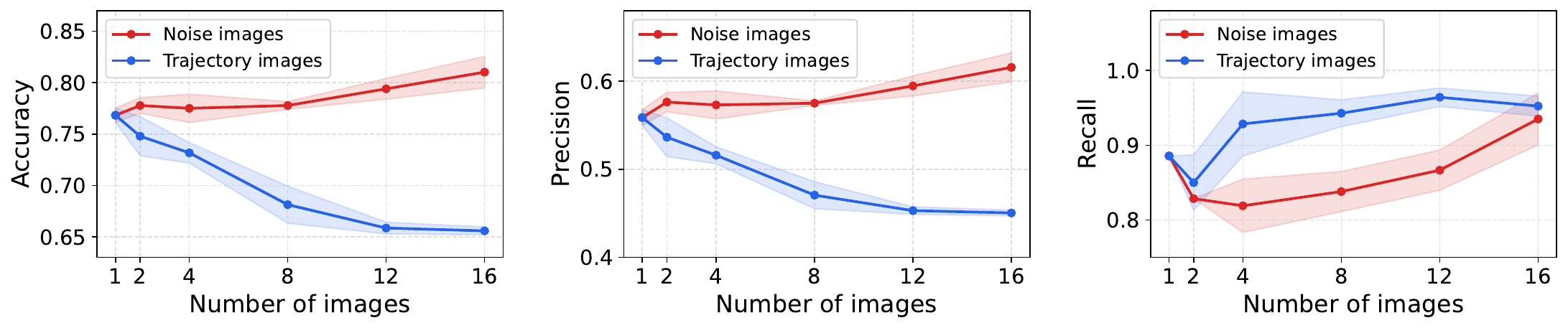}
  \caption{Accuracy, precision and recall against the number of images in one forward pass, with the decisive screenshot always present. Padding uses other screenshots of the same trajectory (\emph{blue}) or noise images of equal token budget (\emph{red}).}
  \label{fig:multiimg_pra}
\end{figure}

%% file: src/appendix/adaptive.tex
\section{Adaptive Compute}
\label{app:adaptive}

\input{src/figures/fig_round_adapt}

SeekJudge spends more compute on longer trajectories. Figure~\ref{fig:round_adapt} groups trajectories into five equal-size bins by step count and reports the mean number of tool calls that the Seek agent issues in each. Tool calls grow slowly with trajectory length. One round may issue several questions, so the longest bin averages more calls than the round cap of $8$.

%% file: src/figures/fig_round_adapt.tex
\begin{figure}[t]
  \centering
  \includegraphics[width=0.5\linewidth]{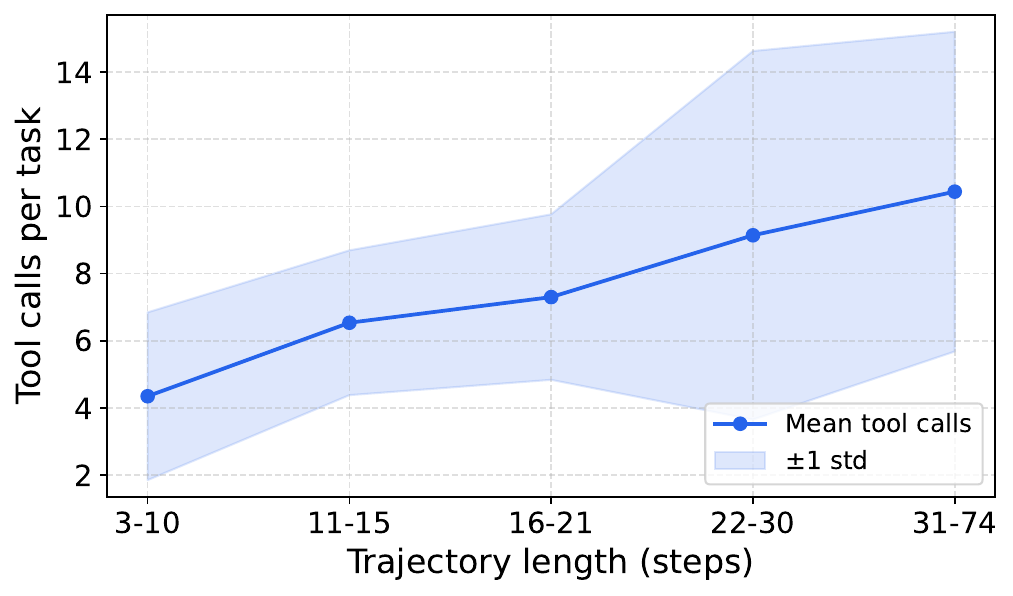}
  \caption{Mean number of tool calls per case against trajectory length, with a $\pm1$ standard deviation band around the mean.}
  \label{fig:round_adapt}
\end{figure}

%% file: src/appendix/leakage.tex
\section{Leakage Control}
\label{app:leakage}

Information from an evaluation target can leak back into SeekJudge at three points, the distillation training data, the prompt search, and the score regression. We control each in turn.

\subsection{Distillation Data and Prompt Search}
\label{app:leakage:prompt}

\paragraph{Training data.} The distillation set that trains SeekJudge shares no task goal with any offline benchmark or online RL environment, so the model is never trained on a case that it is \mbox{later evaluated on}.

\paragraph{Prompt search.} A prompt search can overfit the set that it is scored against. We therefore score both searches on fixed anchor sets and never on benchmark ground truth. The criteria prompt is searched against about $20$ human-annotated trajectories that anchor what counts as success. The seek prompt is searched against the progress descriptions that Codex with GPT-5.5 writes for about $1$K trajectories, and these descriptions guide what to ask and when to stop. Claude Code runs both searches, reading the failures of each case and revising the prompt. All remaining supervision comes from teacher-model distillation.

\input{src/tables/tab_leakage}

\subsection{Score Regression}
\label{app:leakage:fit}

\paragraph{Score regression.} The regressor that maps the seven dimension scores to a trajectory score is an XGBoost model of $200$ trees of depth $2$, with a learning rate of $0.03$, an L2 penalty of $200$ and a minimum child weight of $10$. Each feature is a dimension score divided by $100$, with the sign flipped for the dimensions that lower success, and every feature carries a monotone constraint, so the score cannot decrease when any dimension improves. The model reads only these seven scalar features. One model and one threshold serve every benchmark. The threshold is the value that maximizes Youden's $J$ on the fitted rows pooled across benchmarks and equals $0.374$ on the probability scale, so SeekJudge never selects an operating point per benchmark, and each benchmark receives the same total sample weight in the fit, so a large benchmark does not dominate a small one.

\paragraph{Two fits.} Unless noted otherwise the whole paper uses only two fits of this regressor. The first serves offline evaluation, and we apply it to every offline benchmark, to the trained SeekJudge-9B and the untrained Qwen3-VL-8B backbone alike. The second serves RL. GRPO compares rollouts inside a group that shares one goal, so a dimension that the goal fixes on its own, such as goal infeasibility, takes the same value across the whole group and cannot separate a stronger trajectory from a weaker one. The second fit therefore drops the two goal-level dimensions and reads only the other five dimension scores, with no task identity. We call it the RL calibrator. It is fit once, before training, with the hyperparameters above, on about $1$K rollouts in total of the untrained base policy over the tasks of the three domains, labeled by the environment's rule-based evaluation, and is then frozen. With five features and depth-$2$ trees against about $1$K samples, it has little capacity to store a task-specific solution. GRPO also normalizes the reward within a group, so an offset in $S^{\mathrm{traj}}$ cancels, a change of its scale only reweights the trajectory term against the step term, and the policy sees the relative order of the rollouts of one goal rather than their absolute scores. Appendix~\ref{app:rlfit} replaces this calibrator with the offline fit, which sees no rollout of the RL policy.

\paragraph{Adaptation.} The seven dimension scores are fixed by the prompts, and the mapping from these scores to a verdict is the only part of the pipeline that differs between the two fits. On UI-TARS-1.5-7B Impress, RL with the offline fit in place of the RL calibrator still surpasses rule-based reward (Appendix~\ref{app:rlfit}).

To probe the residual risk we vary how the regressor is fit and read every fit on all three benchmarks, in Table~\ref{tab:leakage}. The setting reported throughout the paper fits jointly on all three, so every reported number reads a benchmark that its fit has seen. Against it we place two kinds of held-out fit. The first is a five-fold cross-fit over the same three benchmarks. Every trajectory is scored by a fold model that never trained on it, and the threshold of each fold comes from that fold's own training rows. The second is three single-benchmark fits. Each produces a diagonal cell that fits and evaluates on the same benchmark and off-diagonal cells that evaluate on benchmarks absent from the fit. In a diagonal cell the regressor has seen the target distribution, so this cell is the highest-risk reading. The cross-fit removes that exposure while keeping the fitting distribution, and an off-diagonal cell removes it while also shifting the fitting distribution to other benchmarks and shrinking the~fitting~set.

Two readings of Table~\ref{tab:leakage} bound the risk. First, the diagonal fits carry the highest risk, yet none of them leads its column by a meaningful margin. On AgentRewardBench the clean transfer fit on CUAStepBench even slightly exceeds it, and on the other two benchmarks it exceeds the nearest off-diagonal transfer by at most half a point. Second, the cross-fit, which never scores a trajectory with a model that trained on it, reaches $75.9$, $75.8$ and $84.6$, within $0.7$ points of the joint fit on AgentRewardBench and OmniGUIRewardBench and $1.6$ points above it on CUAStepBench, so held-out scoring lowers the reported numbers by under one point at most. The single-benchmark transfers add cross-benchmark shift and a smaller fitting set on top of held-out scoring, and even under every fit that excludes the target benchmark, SeekJudge-9B still reaches $75.7$ on AgentRewardBench, $70.1$ on CUAStepBench and $83.8$ on OmniGUIRewardBench, above every baseline on the first two except the agentic Codex run on CUAStepBench ($71.7$ for World-State-Model and $68.5$ for CUAJudge) and equal to CUAJudge on the third ($83.8$, Tables~\ref{tab:offline_cuastepbench} and~\ref{tab:offline_traj}).

%% file: src/tables/tab_leakage.tex
\begin{table}[t]
  \centering
  \caption{Leakage probe for the score regression, reporting trajectory-level F1 (\%) of the trained SeekJudge-9B backbone. Each row is a regression fit on a different set of benchmarks and then read on all three. Row~1 is the joint fit on all three benchmarks that we use throughout the paper (\colorbox{grpyellow}{light yellow}, small overfitting risk). Row~2 is a five-fold cross-fit over the same three benchmarks, with every trajectory scored by a fold model that never trained on it and the threshold set on that fold's own training rows. Rows~3--5 each fit on a single benchmark. A diagonal cell fits and evaluates on the same benchmark (\colorbox{grpred}{light red}, highest overfitting risk), while an off-diagonal cell evaluates on a benchmark absent from its fit (\colorbox{grpgreen}{light green}, lowest overfitting risk).}
  \label{tab:leakage}
  \begin{tabular}{l ccc}
    \toprule
    & \multicolumn{3}{c}{Evaluated on (trajectory F1)} \\
    \cmidrule(lr){2-4}
    Regression fit on & AgentRewardBench & CUAStepBench & OmniGUIRewardBench \\
    \midrule
    All three (ours)      & \cellcolor{grpyellow}76.6 & \cellcolor{grpyellow}74.2 & \cellcolor{grpyellow}85.1 \\
    Five-fold cross-fit   & \cellcolor{grpgreen}75.9 & \cellcolor{grpgreen}75.8 & \cellcolor{grpgreen}84.6 \\
    \midrule
    AgentRewardBench      & \cellcolor{grpred}76.7 & \cellcolor{grpgreen}70.1 & \cellcolor{grpgreen}83.8 \\
    CUAStepBench                & \cellcolor{grpgreen}76.8 & \cellcolor{grpred}74.6 & \cellcolor{grpgreen}85.0 \\
    OmniGUIRewardBench    & \cellcolor{grpgreen}75.7 & \cellcolor{grpgreen}74.1 & \cellcolor{grpred}85.3 \\
    \bottomrule
  \end{tabular}
\end{table}

%% file: src/appendix/rlfit.tex
\section{RL Reward with the Offline Fit}
\label{app:rlfit}

The RL calibrator of Appendix~\ref{app:leakage:fit} is fit on rollouts from the RL environments. To isolate the judge from this fit, we train UI-TARS-1.5-7B on Impress with the offline fit in place of the RL calibrator and keep everything else of the continuous score with the step term unchanged. The offline fit sees no rollout of the RL policy.

\begin{table}[t]
  \centering
  \caption{Rule-based reward and SeekJudge reward (continuous score with the step term) under the two fits of the score regression on UI-TARS-1.5-7B Impress, success rates in \%. The Rule-based and RL calibrator rows repeat the corresponding rows of Table~\ref{tab:reward_granularity}.}
  \label{tab:rlfit}
  \setlength{\tabcolsep}{4pt}
  \begin{tabular}{llcc}
    \toprule
    Reward & Fit & Train & Test \\
    \midrule
    Rule-based & --             & \textbf{43.00} & 30.43 \\
    SeekJudge  & RL calibrator  & 30.06 & \textbf{38.84} \\
    SeekJudge  & Offline fit    & 30.17 & 36.52 \\
    \bottomrule
  \end{tabular}
\end{table}

Table~\ref{tab:rlfit} gives the result. With the offline fit, SeekJudge reaches $36.52$ test success, $6.1$ points above rule-based reward, and the RL calibrator raises it further to $38.84$. In this setting, SeekJudge thus surpasses rule-based reward without the RL calibrator.

%% file: src/appendix/reward_hacking.tex
\section{Reward Hacking during RL}
\label{app:reward_hacking}

During each RL run, every training rollout receives two verdicts, one from the reward model that the policy optimizes and one from the rule-based evaluation of OSWorld, which is also the evaluation used at test time. We write $J$ for the pass rate under the reward model and $G$ for the pass rate under the rule-based evaluation, both smoothed by an exponential moving average with weight $0.95$. When the policy learns behavior that the reward model accepts and the rule-based evaluation rejects, $J$ rises faster than $G$. Figure~\ref{fig:reward_hacking} plots both pass rates for each reward model and application on the Qwen3VL-8B backbone, and Table~\ref{tab:reward_hacking} reports their changes over the $75$ training steps. Each panel uses the same runs as Table~\ref{tab:rl_main}. For CUAJudge we recompute $J$ from the verdicts and rewards in the training log.

\input{src/figures/fig_reward_hacking}

\input{src/tables/tab_reward_hacking}

The column $\Delta J - \Delta G$ of Table~\ref{tab:reward_hacking} is the part of the rise in $J$ that the rule-based evaluation does not confirm. Averaged over the three applications, it is $4.2$ points for SeekJudge, $25.0$ points for CUAJudge, and $2.9$ points for OS-Themis. The rule-based reward optimizes the rule-based evaluation directly, so its value is $0$. SeekJudge raises $J$ by $13.7$ points and $G$ by $9.5$ points, while CUAJudge raises $J$ by $30.1$ points and $G$ by $5.1$ points. OS-Themis stays below the rule-based evaluation throughout, with a gap $J-G$ of $-16.9$ points at the start and $-13.9$ points at the end, and raises $J$ by $7.9$ points. Part of the gain under SeekJudge is not confirmed by the rule-based evaluation, and this part is smaller than under CUAJudge.

%% file: src/figures/fig_reward_hacking.tex
\begin{figure}[t]
  \centering
  \includegraphics[width=\linewidth]{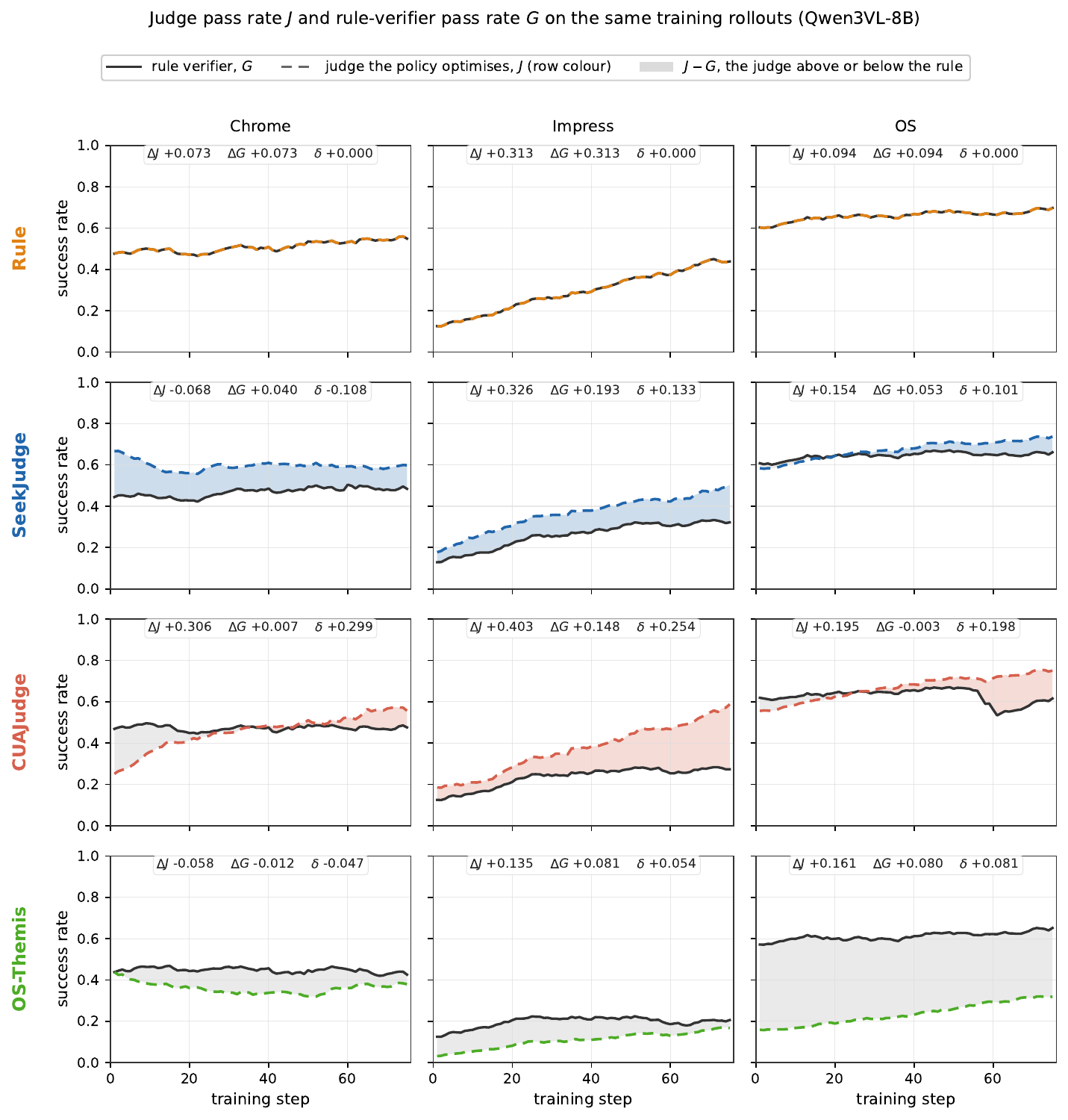}
  \caption{Pass rates on the same training rollouts during RL on the Qwen3VL-8B backbone. The black line is the rule-based evaluation $G$, the dashed line is the reward model $J$, and the shading is the gap between them.}
  \label{fig:reward_hacking}
\end{figure}

%% file: src/tables/tab_reward_hacking.tex
\begin{table}[t]
  \centering
  \caption{Change of the pass rates over the $75$ RL training steps on the Qwen3VL-8B backbone, in points and averaged over Chrome, Impress, and OS. $J$ is the pass rate under the reward model and $G$ is the pass rate under the rule-based evaluation, both on the same training rollouts. $\Delta J - \Delta G$ is the part of the rise in $J$ that the rule-based evaluation does not confirm.}
  \label{tab:reward_hacking}
  \small
  \setlength{\tabcolsep}{6pt}
  \begin{tabular}{lccccc}
    \toprule
    Reward & $\Delta J$ & $\Delta G$ & $\bm{\Delta J - \Delta G}$ & $J-G$ at start & $J-G$ at end \\
    \midrule
    Rule-based & $+16.0$ & $+16.0$ & $\mathbf{0.0}$ & $0.0$ & $0.0$ \\
    SeekJudge & $+13.7$ & $+9.5$ & $\mathbf{+4.2}$ & $+8.2$ & $+12.4$ \\
    CUAJudge & $+30.1$ & $+5.1$ & $\mathbf{+25.0}$ & $-7.4$ & $+17.7$ \\
    OS-Themis & $+7.9$ & $+5.0$ & $\mathbf{+2.9}$ & $-16.9$ & $-13.9$ \\
    \bottomrule
  \end{tabular}
\end{table}

%% file: src/appendix/cost.tex
\section{Inference Cost Model}
\label{app:cost}

This section gives the full cost model behind the cost comparison of Section~\ref{sec:cost}. Every judge's per-case cost is the sum of an open-source token bill and a closed-source API bill, measured on the $43$ cases shared by all four judges.

\paragraph{Token billing.} We bill the self-deployed models by tokens rather than by measured GPU time. GPU time depends on how well each framework parallelizes its requests, so it mixes the concurrency engineering of a serial framework such as OS-Themis into that framework's reported cost.

\paragraph{Token prices.} Open-source agents are billed at the public serving rate of a Qwen3.5-9B or Qwen3-VL-8B class model. This rate is \$$0.04$ per million input tokens and \$$0.20$ per million output tokens.\footnote{\url{https://deepinfra.com/blog/qwen3-5-9b-api-benchmarks}} These prices cover the agents of SeekJudge and the Selector of OS-Themis. They also cover the key-image identification of CUAJudge, because we run it on an open-source model as in its official low-cost setting. Closed-source calls are billed at their provider's published price. The GPT-5-mini calls of CUAJudge are charged at the official OpenAI rate of \$$0.25$ per million input tokens and \$$2.00$ per million output tokens, and the agentic Codex baseline is charged at the GPT-5.5 API spend recorded for each run.

\paragraph{Prefix caching.} We apply one uniform rule to every judge. A self-deployed small model serves a narrow, repetitive workload, so its requests within one case share a long common prefix that the KV cache returns at almost no cost. We therefore treat the prefix that a request shares with an earlier request in the same case as a cache hit, and bill unique input and all output in full. For the Seek agent in SeekJudge and the Selector in OS-Themis, whose input is dominated by one long, reused context, this counts input tokens only for the single longest request of the case and treats the input of every other request as a cache hit at zero cost. CUAJudge issues a single turn per case and has no repeated prefix to reuse, so the same rule gives it no discount. The cache covers only input encoding, so output tokens are always charged in full.

\paragraph{Validation on deployment logs.} We verify this accounting on isolated per-agent logs from our sglang deployment. Within one case, each turn reports a cached-token count equal to the full token count of the preceding request, so the entire prior context is served from the prefix cache and only the marginal new tokens are re-encoded. Summed over the case, the freshly encoded input therefore equals the single longest request, and the accounting above charges exactly this request.

%% file: src/appendix/data.tex
\section{Composition of the Distillation Data}
\label{app:data}

Table~\ref{tab:data_comp} breaks the $10{,}905$ distillation training tasks down by environment. The Android trajectories come from GUI-Odyssey \citep{lu2025guiodyssey}, Mobile-R1 \citep{gu2026mobiler1} and AndroidControl \citep{androidcontrol2024}, the Windows, macOS and Ubuntu trajectories from AgentNet \citep{wang2025opencua}, and the web trajectories from WebArena \citep{zhou2023webarena}, VisualWebArena \citep{koh2024vwa}, WorkArena \citep{drouin2024workarena}, AssistantBench \citep{yoran2024assistantbench} and Online-Mind2Web \citep{xue2025an}. The tasks cover $1{,}473$ distinct applications after merging different spellings of one application, with a long tail. Google Chrome, the most frequent named application, appears in $1{,}569$ tasks, and on Android the most frequent application accounts for under $4$\% of that environment. Figure~\ref{fig:env_app_train} shows the environment split together with the three most frequent applications \mbox{of each environment}.

\input{src/figures/fig_env_app_train}

\input{src/tables/tab_data_comp}

%% file: src/figures/fig_env_app_train.tex
\begin{figure}[t]
  \centering
  \includegraphics[width=\linewidth]{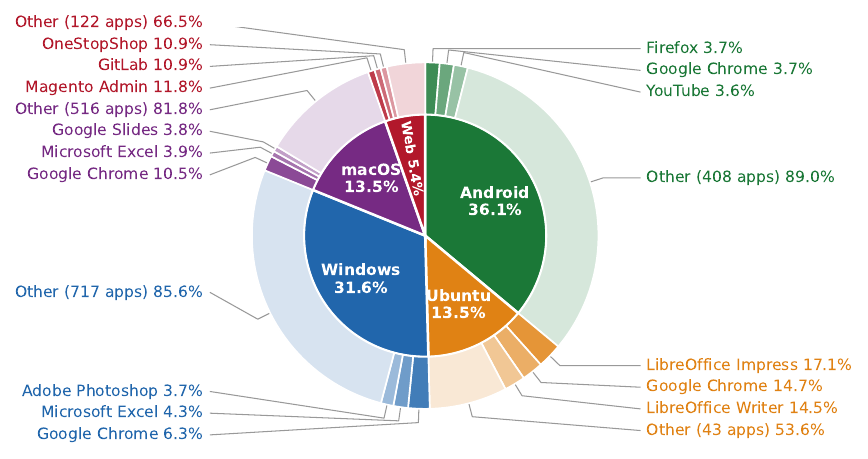}
  \caption{Environments and applications of the distillation trajectories. The inner pie splits the trajectories by environment, and the outer ring splits each environment into its three most frequent applications and Other. A trajectory that uses $k$ applications counts $1/k$ toward each. Each label gives the share of an application within its environment.}
  \label{fig:env_app_train}
\end{figure}

%% file: src/tables/tab_data_comp.tex
\begin{table}[t]
  \centering
  \caption{Composition of the distillation training set by environment. Proportion is over all trajectories, and Apps counts distinct applications, or distinct sites for Web.}
  \label{tab:data_comp}
  \setlength{\tabcolsep}{6pt}
  \begin{tabular}{lcccc}
    \toprule
    Environment & Trajectories & Proportion (\%) & Distinct tasks & Apps \\
    \midrule
    Android & $4{,}000$ & $36.1$ & $4{,}000$ & $411$ \\
    Windows & $3{,}503$ & $31.6$ & $3{,}503$ & $720$ \\
    Ubuntu  & $1{,}500$ & $13.5$ & $1{,}500$ & $46$ \\
    macOS   & $1{,}497$ & $13.5$ & $1{,}497$ & $519$ \\
    Web     & $594$     & $5.4$  & $405$     & $125$ \\
    \midrule
    Total   & $11{,}094$ & $100.0$ & $10{,}905$ & $1{,}473$ \\
    \bottomrule
  \end{tabular}
\end{table}

%% file: src/appendix/bench.tex
\section{Human Curation of CUAStepBench}
\label{app:bench}

Human curation starts from over $300$ candidate tasks. The same annotator checks every label twice. A Codex run then audits the labels and flags $51$ of them as suspicious, and two annotators review each flagged case. They correct $12$ labels, $4.3$\% of the final benchmark, and discard the $7$ cases that remain disputed after discussion. The same Codex configuration appears as a judge in Table~\ref{tab:offline_cuastepbench}.
We further prune cases whose error cause repeats one already covered and balance the remaining trajectories across error types. The cost of running multi-round judges under every configuration that we evaluate sets the final size at $278$ tasks.
An independent expert who took no part in labeling or curation judges all $278$ trajectories blind, from the same instruction, screenshots and actions that a judge receives. Against the final labels, the expert reaches $80.9$\% accuracy, $82.2$ trajectory F1
, listed as Human expert in Table~\ref{tab:offline_cuastepbench}. Each final label, unlike this single blind pass, went through the double check and, when flagged, the two-annotator review described above.

Two properties of this construction matter for the comparison. The uncertainty filter keeps the cases that leave the teacher pipeline unsure and discards the cases that it judges with confidence, whether rightly or wrongly, and every judge in Table~\ref{tab:offline_cuastepbench} is scored on the same retained set. The Codex audit only flags labels for review, and a label changes only after two annotators confirm the change. AgentRewardBench, which is human-labeled and whose cases we did not select, gives a second reading. There SeekJudge-9B leads CUAJudge by $14.0$ points, against $5.7$ points on CUAStepBench (Tables~\ref{tab:offline_cuastepbench} and~\ref{tab:offline_traj}).

Figure~\ref{fig:env_app_bench} breaks the $278$ CUAStepBench trajectories down by environment and application. Android contributes $98$ trajectories, Windows $75$, Ubuntu $43$, macOS $33$ and Web $29$, over $177$ distinct applications. These proportions are close to those of the distillation data in Figure~\ref{fig:env_app_train}, except that the Web share of $10.4$\% is about twice the $5.4$\% of the distillation data.

\input{src/figures/fig_env_app_bench}

The disagreement study of Figure~\ref{fig:multiimg}(a) uses the same kind of review. A Codex judge runs over every trajectory of OmniGUIRewardBench, and a human re-checks each of its $71$ disagreements with the rule without knowing which verdict comes from the rule and which from the judge, agreeing with the judge on $48$.

%% file: src/figures/fig_env_app_bench.tex
\begin{figure}[t]
  \centering
  \includegraphics[width=\linewidth]{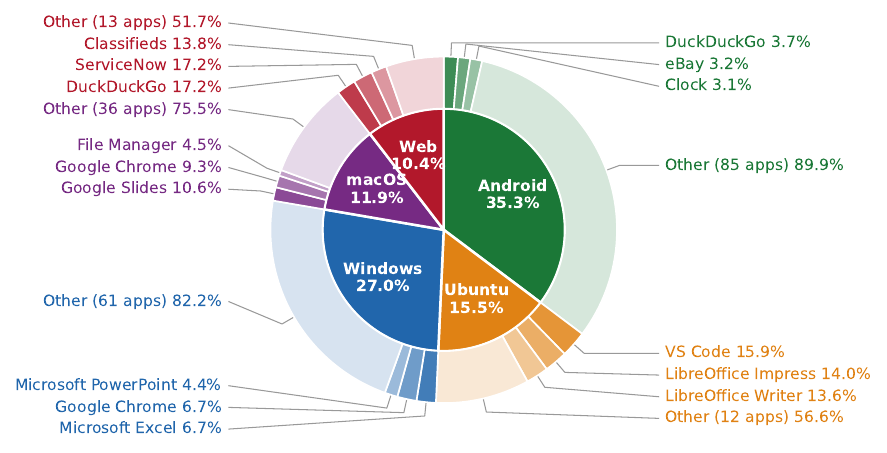}
  \caption{Environments and applications of the $278$ CUAStepBench trajectories. The inner pie splits the trajectories by environment, and the outer ring splits each environment into its three most frequent applications and Other.}
  \label{fig:env_app_bench}
\end{figure}

%% file: src/appendix/training.tex
\section{Training Details}
\label{app:setup}

\subsection{Supervised Fine-Tuning of SeekJudge-9B}
\label{app:sft}

We obtain SeekJudge-9B by full-parameter supervised fine-tuning of the language model of Qwen3.5-9B, keeping the vision encoder and the vision-language aligner frozen. The training set holds $45{,}416$ samples and about $151$M tokens per epoch, and it mixes trajectory-level judging targets, per-step analysis targets, and tool-query targets. We train for $2$ epochs with ms-swift, using AdamW with $\beta=(0.9, 0.95)$, a weight decay of $0.01$, a learning rate of $1\mathrm{e}{-}5$ with a cosine schedule and a warmup ratio of $0.05$, gradient clipping at $1.0$, a global batch size of $32$ sequences, and a maximum sequence length of $32$K tokens. Training runs on $8\times$ RTX A6000 GPUs with DeepSpeed ZeRO-3 and CPU offloading and takes about $41$ hours.

\subsection{Reinforcement Learning Configuration}
\label{app:training}

This subsection lists the RL hyperparameters and the evaluation protocol that Section~\ref{sec:training} summarizes. We optimize every actor with GRPO and train one policy per domain across the two backbones UI-TARS-1.5-7B and Qwen3-VL-8B, keeping the domains separate so that each number reflects a small model specialized to a single domain. Every run uses a train batch size of $8$ with a group size of $8$ rollouts per prompt, a learning rate of $1\mathrm{e}{-}6$, and an actor image-history length of $2$, and runs for $75$ training steps on $8\times$A100 GPUs in roughly one day. The GRPO objective uses no KL penalty, clipping ratios of $0.2$ and $0.28$ for the lower and upper bounds, and token-mean loss aggregation. Rollouts sample at temperature $1.0$, and test evaluation decodes greedily. The Chrome, Impress, and OS domains hold $23$, $24$, and $12$ training tasks and $23$, $23$, and $12$ test tasks.

For test evaluation we run the policy every $15$ training steps, repeating the pass three times over a held-out test set at each checkpoint. The value reported in the main results is the mean over all test evaluations collected during training. Averaging over checkpoints reduces the fluctuation of any single checkpoint. The training success rate is measured by rule-based evaluation on the training tasks and reported as an exponential moving average over training steps. Each rollout episode is capped at $15$ environment steps during training and $25$ during test evaluation, following ACuRL \citep{cuajudge2025}, with a maximum prompt length of $13$K tokens and a response length~of~$512$~tokens.

\input{src/figures/fig_rl_val_curve}

Figure~\ref{fig:rl_val_curve} plots the test success of every reward at the five test evaluations, taken every $15$ of the $75$ training steps, averaged over the two backbones and three domains. SeekJudge stays above every other reward at all five training steps. Rule-based reward rises steadily over training, while CUAJudge and OS-Themis end below their value at step $15$.

\subsection{Trajectory Reward from Dimension and Step Scores}
\label{app:reward}

This subsection gives the implementation of the trajectory reward. The reward server returns the trajectory score $S^{\mathrm{traj}}_i$ from the regressor and the step score $S^{\mathrm{step}}_{i,k}$ that the rubric assigns to every step, and Eq.~\ref{eq:reward} combines them into the trajectory reward $R_i$. It averages the step scores instead of summing them, keeping this term on a common scale so a longer rollout earns no extra reward for taking more steps. The weight $w_{\mathrm{step}}=2$ was set once before training and never tuned, and we tried no other value. A group holds the $G=8$ rollouts of one goal, and the GRPO advantage of step $k$ of rollout $i$ normalizes $R_i$ over the $G$ trajectory rewards,
\begin{equation}
\hat{A}_{i,k} = \frac{R_i - \operatorname{mean}(\{R_j\}_{j=1}^{G})}{\operatorname{std}(\{R_j\}_{j=1}^{G})}.
\label{eq:adv}
\end{equation}
The right side carries no step index $k$. Every decision step of the rollout shares this one advantage, and the step scores enter it only through their mean in Eq.~\ref{eq:reward}. We leave per-step advantage estimation to future work. CUAJudge and OS-Themis enter the same pipeline through their native binary verdicts and receive no such calibration.

%% file: src/figures/fig_rl_val_curve.tex
\begin{figure}[t]
  \centering
  \includegraphics[width=0.5\linewidth]{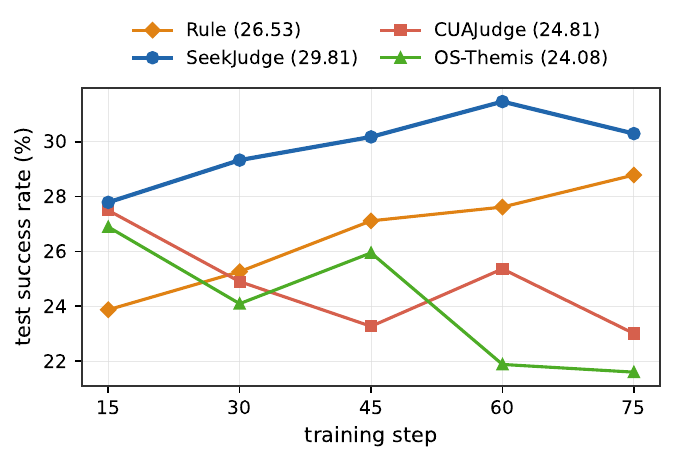}
  \caption{Test success during RL at training steps $15$, $30$, $45$, $60$ and $75$, averaged over the six backbone-domain settings of Table~\ref{tab:rl_main}. The value in the legend is the mean of each curve over these five steps, which equals the average test success of that reward in Table~\ref{tab:rl_main}.}
  \label{fig:rl_val_curve}
\end{figure}

%% file: src/appendix/stepclass.tex
\section{Step Classes and Step Scores}
\label{app:stepclass}

Table~\ref{tab:stepclass} defines the nine step classes that the step prompt assigns, one class per step, and lists the step score $S^{\mathrm{step}}$ that each class contributes to the RL reward of Eq.~\ref{eq:reward}. For offline evaluation, the six classes marked correct count as correct steps and the other three count as erroneous steps. This split approximates the criterion of the human step labels in CUAStepBench, which mark a step as erroneous only when it harms the final result or a later step uses its wrong result. A detour or a missed warning that the agent later recovers from counts as correct under this criterion.

\input{src/tables/tab_stepclass}

%% file: src/tables/tab_stepclass.tex
\begin{table}[t]
  \centering
  \caption{The nine step classes, their step scores $S^{\mathrm{step}}$ in the RL reward, and the classes counted as correct steps in offline evaluation (Correct). The other three classes are the erroneous steps.}
  \label{tab:stepclass}
  \small
  \setlength{\tabcolsep}{4pt}
  \begin{tabular}{l p{0.54\linewidth} c c}
    \toprule
    Class & Definition & $S^{\mathrm{step}}$ & Correct \\
    \midrule
    \texttt{milestone} & Completes a sub-goal of the task by producing its end-state~result. & $+0.15$ & \cmark \\
    \texttt{correct\_decision} & Picks the right one of at least two visible options when the choice needs icon recognition or application knowledge rather than text matching. & $+0.10$ & \cmark \\
    \texttt{error\_correction} & Reverses an earlier wrong action. & $+0.05$ & \cmark \\
    \texttt{neutral} & A step on the correct workflow that fits no other class. & $0$ & \cmark \\
    \texttt{off\_path} & A step outside the correct workflow, such as a detour or a step of a wrong strategy. & $-0.05$ & \cmark \\
    \texttt{missed\_correction} & Continues past a clear on-screen sign of a problem, such as an error message or an empty search result. & $-0.05$ & \cmark \\
    \texttt{wrong\_action} & Performs a wrong operation whose result a later step uses or that gives a wrong final result. & $-0.15$ & \xmark \\
    \texttt{meaningless} & Repeats an action whose identical result has already been observed, and no later step needs it. & $-0.05$ & \xmark \\
    \texttt{false\_claim} & Claims completion of an incomplete task when no earlier wrong action caused the failure. & $-0.15$ & \xmark \\
    \bottomrule
  \end{tabular}
\end{table}

%% file: src/appendix/prompt.tex
\section{Prompt Construction}
\label{app:context}

The judging process uses three prompts, the seek prompt that explores the trajectory, the overall-trajectory prompt that scores the trajectory, and the step prompt that scores each step. We call the last two together the criteria prompt. Separating exploration from scoring lets the step prompt be reused in other frameworks unchanged.

The seek prompt drives exploration. At each round the agent either calls a tool to gather more evidence or gives its conclusion. Giving a conclusion ends the phase, and otherwise the agent keeps exploring until it reaches the round limit, and then it is forced to stop and conclude.

Once the conclusion is fixed, the criteria prompt produces the detailed evaluation. It splits into two independent scoring prompts, the overall-trajectory prompt that assigns the seven dimension scores and the step prompt that assigns a step class to each step, each under its own system prompt. The step prompt depends only on the trajectory representation rather than on the exploration context, so it attaches to any framework that lacks native step-level judgment, including CUAJudge and OS-Themis, and yields the step labels that we report for those baselines.

An untrained model, such as the SeekJudge Qwen3-VL-8B backbone or CUAJudge, runs both phases in full at inference. For the trained SeekJudge model we merge the second phase into the first during distillation, so the training data has the model output the detailed evaluation directly at the end of exploration rather than under a separate scoring prompt.

We excerpt the two criteria prompts below. The excerpts keep the structure, the definitions and the decision rules of each prompt and replace its examples and repeated clarifications with [\ldots].

\paragraph{Overall-trajectory prompt.}
\begin{promptexcerpt}
\textbf{Overall Trajectory Evaluation: 7-Dimension Scoring (0-100)}\\
You are evaluating an agent's GUI task trajectory. Score each dimension 0-100 based on the goal, actions, and observable results. CRITICAL DISTINCTION: You must distinguish between the agent's PERFORMANCE and the ENVIRONMENT's PROPERTIES. [\ldots]

\textbf{1. goal\_infeasibility.} The likelihood the goal is INHERENTLY impossible in this environment, REGARDLESS of agent behavior. 0 = Goal is clearly feasible. 50 = Uncertain. 100 = Goal is impossible, required functionality/data provably absent. [\ldots] If 1 of N sub-tasks targets a non-existent object, score proportionally. Agent failure due to mistakes does NOT increase infeasibility. [\ldots]

\textbf{2. goal\_ambiguity.} How ambiguous or underspecified the goal text is. 0 = Perfectly unambiguous; exactly one correct outcome. 50 = Some ambiguity. 100 = Extremely vague; many valid interpretations. This is about the GOAL TEXT itself, not the agent's understanding~of~it.~[\ldots]

\textbf{3. exploration\_degree.} How appropriately the agent explored relative to the task's needs. First determine if the task REQUIRES exploration. [\ldots] Routine UI navigation (clicking tabs, opening menus, selecting options along a known path) is NOT~exploration.~[\ldots]

\textbf{4. path\_redundancy.} How much unnecessary redundancy exists in the path. 0 = Most direct path possible. 50 = About half the steps were unnecessary. 100 = Path dominated by wasted steps. [\ldots] A long trajectory is NOT necessarily redundant if most steps serve~the~goal.~[\ldots]

\textbf{5. positive\_evidence\_reliability.} How reliable the POSITIVE evidence is that the goal was completed CORRECTLY. [\ldots] An action on the wrong target produces ZERO positive evidence for that sub-task, even if the action itself was executed flawlessly. [\ldots] An UNVERIFIED\_CLAIM is NOT positive evidence by itself. [\ldots] Use prior analysis (tool call results) as ground truth. [\ldots]

\textbf{6. negative\_evidence\_reliability.} How reliable the NEGATIVE evidence is that the goal was NOT completed correctly. [\ldots] Negative evidence is about the OUTCOME, not the PROCESS. [\ldots]

\textbf{7. completion\_progress.} How much of the goal the agent CORRECTLY completed. [\ldots] A sub-task performed on the wrong target does NOT count as progress. [\ldots]

\textbf{Analysis Process.} Step 1: Goal Decomposition. Step 2: Sub-task Verification. Step 3: Error Detection. Step 4: Independent Scoring. Step 5: Consistency Check. [\ldots] You MUST output exactly 7 labels. [\ldots] All values must be integers between 0 and 100.
\end{promptexcerpt}

\paragraph{Step prompt.}
\begin{promptexcerpt}
\textbf{Step-Level Analysis: 9-Category Per-Transition Classification}\\
Before outputting results, follow this analysis process:

\textbf{Step A: Goal Decomposition.} List each sub-goal the agent must achieve. Each sub-goal should correspond to a user-visible outcome that directly satisfies part of the goal, NOT an intermediate implementation step. [\ldots] For each sub-goal, identify its end-state deliverable. [\ldots]

\textbf{Step B: Completion Check.} For each sub-goal, determine: was it actually completed, not just attempted? [\ldots] The agent claiming "done" does NOT prove completion. Verify against the~actual~state.~[\ldots]

\textbf{Step C: Correct Approach Identification.} Determine what the correct workflow for this task would be. [\ldots] For each actual transition, determine whether it is on-path or off-path. [\ldots] If the agent's overall strategy is wrong, then ALL steps executing that wrong strategy are off-path. [\ldots]

\textbf{Step D: Error Analysis.} For each failed sub-goal, trace the cause: 1.~Did the agent perform a concretely wrong operation whose result was consumed by later steps? If yes, that step is a wrong\_action. [\ldots] 2.~After that wrong action, did the agent successfully undo/reverse it? If yes, that corrective step is an error\_correction. 3.~Did the environment produce a clear, observable signal that something was wrong, and the agent continued without addressing it? If yes, that step is a missed\_correction. 4.~If no wrong operation was consumed downstream, but the agent claimed completion when the task was incomplete, that step is a false\_claim. [\ldots]

\textbf{Step E: Achievement Analysis.} 1.~Which step made the sub-goal transition from not-done to done? That step is a milestone. [\ldots] 2.~At any step, did the agent select correctly from $\geq$2 visually present alternatives, where the correct option required non-textual reasoning (icon recognition, indirect navigation, domain knowledge)? That step is a correct\_decision. [\ldots]

\textbf{Step F: Remaining Steps.} 1.~Does this transition REVERSE progress toward a correct state? If yes, off\_path. [\ldots] 2.~Is the action an exact repeat with identical result already observed AND not causally necessary? If yes, meaningless. 3.~Is this step off-path (from Step C)? If yes, off\_path. 4.~Otherwise, neutral.

\textbf{Category Definitions (9 categories, mutually-exclusive).} [\ldots]\\
wrong\_action. Key test: Was the wrong output used downstream, or is this the final step that produces an incorrect result? [\ldots]\\
false\_claim. Key test: Is it the done-report AND is the failure from omission (not from a prior wrong\_action)? [\ldots] Only wrong\_action blocks false\_claim. [\ldots]\\
error\_correction. Key test: Does it explicitly undo a specific~prior~error?\\
missed\_correction. Key test: Did the environment produce an objectively visible signal of a problem, and the agent proceeded without correcting it? [\ldots] Only mark the most~salient~signal.~[\ldots]\\
milestone. Key test: Can I point to which sub-goal was completed? Is this the step that produces the end-state deliverable, not a preparatory step? [\ldots]\\
correct\_decision. Key test: Are there $\geq$2 plausible options visible AND the agent picked right AND a text-only user couldn't make this choice? [\ldots]\\
off\_path. Key test: Would an ideal agent who knows the correct approach have taken this step? If no, off\_path. [\ldots]\\
meaningless. Key test: Was this exact action+result already seen before? [\ldots]\\
neutral. Key test: Is this step on the correct workflow path AND does it not fit any other category?

[\ldots] Each transition gets exactly ONE category. [\ldots] You MUST output exactly one line per transition, covering ALL transitions from Transition 1 to the last transition in the trajectory. [\ldots]
\end{promptexcerpt}